%% file: colm2024_conference.tex
\documentclass{article}
\usepackage{colm2024_conference}

\usepackage{booktabs}
\usepackage{graphicx}
\usepackage{amsmath}
\usepackage{enumitem}
\usepackage{wrapfig}
\usepackage{algorithm}
\usepackage{algpseudocode}
\usepackage{natbib}
\usepackage{makecell}%
\usepackage{booktabs}
\usepackage{array}
\usepackage{amssymb}
\usepackage{amsfonts}
\usepackage{multirow}
\usepackage{verbatim}
\usepackage{caption}
\usepackage{longtable}
\usepackage{supertabular}
\usepackage{CJKutf8}
\usepackage[utf8]{inputenc}
\usepackage[T1]{fontenc}
\usepackage[french,vietnamese,mongolian,greek,english]{babel}
\usepackage{pifont}

\usepackage{enumitem}
\usepackage{tablefootnote}

\usepackage{xspace}
\usepackage{textcomp}
\usepackage{makecell}
\usepackage{lscape} 
\usepackage{siunitx}
\usepackage{listings}
\usepackage{xcolor}
\usepackage{svg}

\definecolor{dt}{gray}{0.7}

\usepackage{pifont}
\usepackage{bbding}
\usepackage{fontawesome}

\usepackage{scrextend}

\usepackage{tgpagella}
\usepackage{latexsym}
\usepackage[T1]{fontenc}
\usepackage[utf8]{inputenc}
\usepackage{microtype}
\definecolor{mydarkblue}{rgb}{0,0.08,0.45}
\definecolor{citecolor}{HTML}{0071BC}
\usepackage{url}
\usepackage{nicefrac}
\usepackage{changepage}
\usepackage{xargs}
\usepackage{wrapfig,lipsum,booktabs}
\usepackage{longtable}
\usepackage{subcaption}
\usepackage{endnotes}
\usepackage{tabularx}

\usepackage{pgfplots}
\usetikzlibrary{pgfplots.groupplots}
\pgfplotsset{compat=1.3}
\usepackage{tikz}
\usetikzlibrary{patterns}

\usepackage[most]{tcolorbox}

\usepackage{hyperref}
\definecolor{darkblue}{rgb}{0, 0, 0.5}
\hypersetup{colorlinks=true, citecolor=darkblue, linkcolor=darkblue, urlcolor=darkblue, bookmarks=true, bookmarksopen=false, bookmarksnumbered=true}

\usepackage[capitalize,noabbrev]{cleveref}

\crefname{section}{\S}{\S\S}
\Crefname{section}{\S}{\S\S}
\crefname{subsection}{\S\S}{\S\S}
\Crefname{subsection}{\S\S}{\S\S}
\crefformat{section}{\S#2#1#3}
\crefformat{subsection}{\S\S#2#1#3}

\crefname{table}{Table}{Tables}
\crefname{figure}{Figure}{Figures}
\crefname{algorithm}{Algorithm}{}
\crefname{equation}{eq.}{}
\crefname{appendix}{Appendix}{}
\crefformat{section}{Section #2#1#3}
\usepackage{multicol}
\usepackage{tcolorbox}
\usepackage{titlesec}
\titleformat*{\section}{\large\bfseries}
\usepackage{array}
\newcolumntype{P}[1]{>{\centering\arraybackslash}p{#1}}
\usepackage{multirow}

\usepackage{adjustbox}
\definecolor{objblue}{RGB}{3,139,221}  
\definecolor{attrred}{RGB}{255,67,67}    
\definecolor{easygreen}{RGB}{0,156,75}  
\definecolor{middleyellow}{RGB}{242,89,34}  
\definecolor{hardred}{RGB}{216,56,58}

\usepackage{colortbl}
\usepackage{array}

\usepackage[most]{tcolorbox}
\definecolor{BoxBackground}{RGB}{240, 240, 240}
\definecolor{BoxFrame}{RGB}{0, 0, 0}
\definecolor{TitleBackground}{RGB}{0, 0, 0}
\definecolor{TitleText}{RGB}{255, 255, 255}

\tcbset{
  academicbox/.style={
    boxsep=5pt,
    left=2pt,
    right=2pt,
    bottom=0.5pt,
    boxrule=0.5pt,
    colback=BoxBackground,
    colframe=BoxFrame,
    colbacktitle=TitleBackground,
    coltitle=TitleText,
    enhanced,
    attach boxed title to top left={yshift=-0.1in,xshift=0.1in},
    boxed title style={boxrule=0pt,colframe=white},
    title={#1},
  }
}
\newtcolorbox{AcademicBox}[1][]{academicbox=#1}

\title{Qwen-Image-Bench: From Generation to Creation \\
in Text-to-Image Evaluation}

\author{
\small
Niantong Li,
Guangzheng Hu,
Weixu Qiao,
Ying Ba,
Qichen Hong,
Shijun Shen,
Jinlin Wang, \\
Fan Zhou,
Jianye Kang,
Xin Shang,
Ziyi He,
Wei Wang,
Dalin Li,
Jiahao Li,
Jie Zhang, \\
Kaiyuan Gao,
Kun Yan,
Lihan Jiang,
Ningyuan Tang,
Shengming Yin,
Tianhe Wu,\\
Xiao Xu,
Xiaoyue Chen,
Yuxiang Chen,
Yan Shu,
Yanran Zhang,
Yilei Chen,\\
Yixian Xu,
Zekai Zhang,
Zhendong Wang,
Zihao Liu,
Zikai Zhou, \\
Hongzhu Shi,
Yi Wang,
Bing Zhao,
Hu Wei,
Lin Qu,
Chenfei Wu\thanks{Corresponding author} \\
\texttt{liniantong.lnt@alibaba-inc.com, fulai.hr@alibaba-inc.com}
}

\begin{document}

\maketitle

\input{content/section_abstract}

\section{Introduction}
\input{content/section_introduction}

\section{Related Work}
\input{content/section_related_work}

\section{Qwen-Image-Bench}
\input{content/section_benchmark}

\section{Experiment}
\input{content/section_experiment}

\section{Conclusion}
\input{content/section_conclusion}

\clearpage

\clearpage
\bibliography{colm2024_conference}
\bibliographystyle{colm2024_conference}

\newpage

\input{content/section_appendix}
\end{document}

%% file: content/section_abstract.tex
\begin{abstract}
Text-to-Image (T2I) generation has evolved from basic image synthesis into a frequently used core capability in professional creative workflows, where simple text-image alignment can no longer satisfy users' pressing demands for faithful real-world reconstruction and genuine creative expression. Existing benchmarks, however, remain anchored in these foundational criteria and do not yet capture the nuanced capabilities that matter in authentic artistic practice, making it difficult to reliably distinguish state-of-the-art T2I models. Moreover, many recent evaluation pipelines rely heavily on a unsupervised multimodal large language model (MLLM) as the sole judge, diverging from professional human standards. To address these gaps, we introduce \textbf{Qwen-Image-Bench}, a creator-centric benchmark co-designed with professional artists and grounded in real-world creation scenarios. Building upon the conventional pillars of \textit{Quality}, \textit{Aesthetics}, and \textit{Text-Image Alignment}, our benchmark enriches evaluation with two application-driven dimensions: \textit{Real-world Fidelity} and \textit{Creative Generation}. Drawing on the staged reasoning inherent in professional artistic workflows, we organize these five pillars into a top-down hierarchical taxonomy that further decomposes into 23 second-level sub-capabilities and 56 third-level verifiable rubrics. To ensure broad coverage, we curate 1{,}000 stratified bilingual prompts balanced across length and language, with each prompt jointly exercising more than four fine-grained facets across multiple pillars. We train a unified judge model (\textbf{Q-Judger}) based on Qwen3.6-27B, supervised by 80 professional annotators from art academies under blind labeling and triple-review protocols, that scores every image across all 56 verifiable facets, producing fine-grained, rubric-grounded, and fully attributable diagnostics rather than a single opaque score. Empirically, Qwen-Image-Bench reliably distinguishes leading T2I models across both Chinese and English prompts, achieving the greatest separation on the two application-driven dimensions of \textit{Real-world Fidelity} and \textit{Creative Generation} where existing benchmarks provide little insight, while also providing a trustworthy optimization signal for production-level T2I development.
\end{abstract}

%% file: content/section_introduction.tex
Text-to-image (T2I) generation has advanced rapidly from producing visually plausible images to supporting photorealistic world reconstruction and rich creative expression \citep{wang2025designdiffusionhighqualitytexttodesignimage,xu2023imagerewardlearningevaluatinghuman,zhang2024texttoimagesynthesisdecadesurvey,openai2023dalle3,wu2025qwenimagetechnicalreport,imageteam2025zimageefficientimagegeneration,nanobananapro2025,wang2025dreamtexthighfidelityscene,stability2024sd3,seedream2025seedream40nextgenerationmultimodal}. Driven by continuous progress in web-scale training \citep{li2024scalabilitydiffusionbasedtexttoimagegeneration}, stronger text encoders \citep{li2024textcraftortextencoderimage,wang2025scalingtextencoderstexttoimage}, and widely adopted techniques such as classifier-free guidance \citep{bradley2024classifierfreeguidancepredictorcorrector}, models based on either latent diffusion \citep{stability2024sd3,blackforestlabs2025flux2,deepmind2025imagen,openai2023dalle3} or autoregressive token modeling \citep{wu2024janusdecouplingvisualencoding,chen2025blip3ofamilyfullyopen,imageteam2025zimageefficientimagegeneration,jiang2025t2ir1reinforcingimagegeneration} have largely saturated basic prompt following. Exemplified by Nano Banana Pro \citep{nanobananapro2025}, which emphasizes real-world knowledge and fine-grained controllability, and GPT Image 1.5 \citep{openai2025chatgptimages}, which delivers notable gains in text rendering and facial fidelity, frontier production-grade models jointly signal a decisive shift in the field. The bottleneck is no longer whether the generated image aligns with the prompt, but whether it achieves high \textit{Real-world Fidelity}, faithfully reconstructing physical, cultural, and factual reality, and enables authentic \textit{Creative Generation} that supports originality, aesthetic intention, and stylistic coherence. These two capabilities now determine a model's practical value in professional creator workflows.

Despite this rapid progress in generation, T2I evaluation has lagged behind. Most established benchmarks still focus on a narrow set of basic criteria such as coarse text-image alignment and generic visual quality, implemented mainly through CLIP similarity and its variants \citep{bakr2023hrsbenchholisticreliablescalable,wu2023humanpreferencescorev2,boomb0om2023text2imagebenchmark}. Even recent efforts that introduce vision-language judges (e.g., TIFA \citep{hu2023tifaaccurateinterpretabletexttoimage} and GenAI-Bench \citep{li2024genaibenchevaluatingimprovingcompositional}) or preference-based evaluators (e.g., ImageReward \citep{xu2023imagerewardlearningevaluatinghuman} and HPS variants \citep{wu2023humanpreferencescorev2}) remain fragmented in scope and largely tied to semantic correctness, lacking coverage of the professional capabilities that truly distinguish frontier commercial models. We identify three key limitations. \textbf{(1) Saturation on basic semantic axes reduces discriminative power.} As modern T2I models approach the ceiling on common prompt-following suites, widely used benchmarks increasingly compress strong models into indistinguishable score bands, and some even exhibit benchmark drift, where scores drift from human judgment as model capabilities evolve \citep{kamath2025geneval2addressingbenchmark,chen2025multimodallanguagemodelstexttoimage}. \textbf{(2) \textit{Real-world Fidelity} remains insufficiently characterized.} Existing suites rarely separate visually plausible outputs from those that genuinely respect physical commonsense, cultural grounding, or factual structure; these gaps are only partially addressed by recent probes on physics, knowledge, and cultural expectations \citep{meng2024phybenchphysicalcommonsensebenchmark,niu2025wiseworldknowledgeinformedsemantic,zhang2025worldgenbenchworldknowledgeintegratedbenchmarkreasoningdriven,nayak2026culturalframesassessingculturalexpectation}. \textbf{(3) \textit{Creative Generation} lacks well-defined evaluation protocols and adequate expert supervision.} Creators place first-order demands on aesthetics, originality, and stylistic coherence during key workflow stages (ideation, styling, and iterative refinement), yet most benchmarks lack an expert-defined, fine-grained taxonomy along these axes. Compounding this, a growing number of evaluation pipelines place uncritical trust in a single multimodal large language model (MLLM) as the judge, and even train downstream reward models directly on its labels. For instance, UniGenBench++ \citep{wang2025unigenbenchunifiedsemanticevaluation} adopts Gemini-2.5 outputs as supervision, so the resulting scores inherit the judge's systematic biases rather than reflecting professional human standards grounded in trained aesthetic intuition and domain expertise.

\begin{figure*}[t]
  \includegraphics[width=\textwidth]{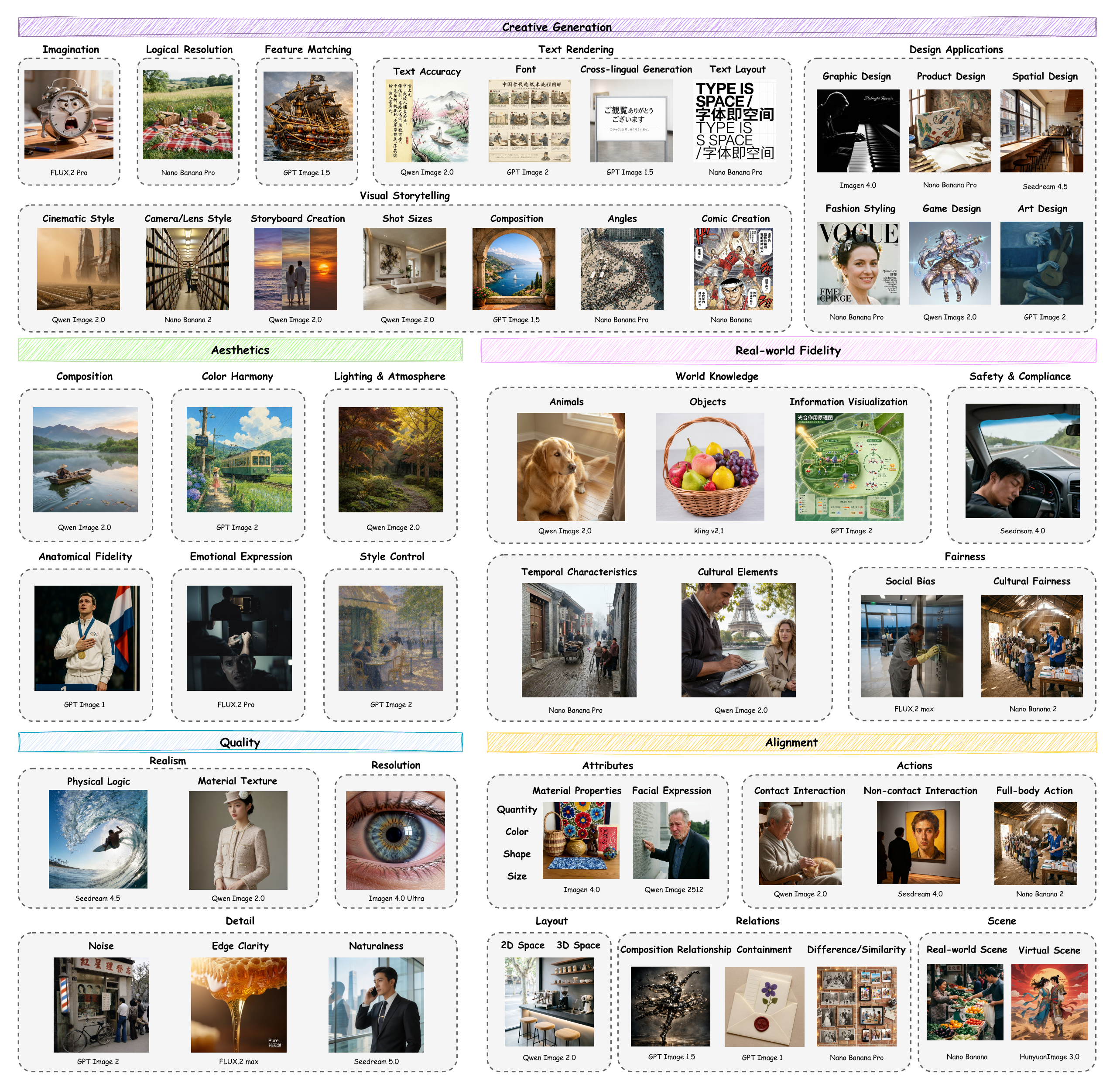}
  \vspace{-7mm}
  \caption{Qwen-Image-Bench Evaluation Dimensions.}
  \vspace{-5mm}
  \label{fig:teaser}
\end{figure*}

To address these gaps, we introduce \textbf{Qwen-Image-Bench}, a \textit{creator-centric} benchmark built around authentic creator demands and co-developed with professional artists. Moving beyond semantic correctness, it jointly evaluates \textit{Real-world Fidelity} and \textit{Creative Generation} alongside the three conventional pillars of \textit{Quality}, \textit{Aesthetics}, and \textit{Text-Image Alignment}. The benchmark highlights dimensions that matter most in real-world creation, such as \textit{World Knowledge}, \textit{Text Rendering}, and \textit{Visual Storytelling}, based on what creators actually care about in their daily workflows. As shown in Fig.~\ref{fig:teaser}, the benchmark comprises 1{,}000 expert-crafted structured prompts organized under a three-level taxonomy, curated through rigorous filtering and stratified sampling to guarantee broad, balanced coverage and evaluation efficiency.

For the taxonomy itself, rather than listing evaluation facets as a flat checklist, we design it top-down along the staged reasoning of professional artistic workflows, specifically mirroring the key phases of \textit{ideation}, \textit{styling}, and \textit{iterative refinement}. Starting from what a creator wants to express at the high level (L1), we drill down into the sub-capabilities they need (L2), and finally into the concrete details a reviewer would inspect (L3). This yields \textbf{5 first-level pillars}, which further decompose into \textbf{23 second-level sub-capabilities} and \textbf{56 third-level evaluation facets}. This decomposition offers several design advantages. Since each pillar is unfolded against concrete artistic sub-processes rather than ad-hoc brainstorming, the resulting taxonomy is more systematic, exhaustive, and less prone to overlap. It also provides a direct anchor for prompt construction: third-level facets act as atomic coverage units, allowing each prompt to jointly exercise multiple facets across pillars, yielding denser and more balanced dimensional coverage than prompt sets built on flat categories. In addition, diagnosis becomes interpretable at the capability level, localizing a model's strengths and weaknesses to specific artistic sub-skills rather than collapsing them into an aggregate score. Among the 56 third-level facets, 28 fall under the two application-driven pillars of \textit{Real-world Fidelity} and \textit{Creative Generation}, covering high-frequency creator scenarios such as world knowledge, design applications, visual storytelling, and text rendering.

Regarding prompt design, many existing benchmarks are dominated by short prompts, with average lengths of 16.43 and 12.65 tokens in HRS-Bench \citep{bakr2023hrsbenchholisticreliablescalable} and T2I-CompBench \citep{huang2025t2icompbenchenhancedcomprehensivebenchmark} respectively. Such short prompts often reflect vague user requests, test only a single narrow aspect, and fail to cover the rich creative intentions that arise in real-world creation. Dense-prompt suites such as DPG-Bench \citep{hu2024ellaequipdiffusionmodels} and UniGenBench++ \citep{wang2025unigenbenchunifiedsemanticevaluation}, while explicitly accounting for prompt complexity, are still typically capped around 80 to 160 tokens and struggle to carry the rich scene context found in real creator requests. To better reflect practical usage, Qwen-Image-Bench deliberately balances 500 long prompts and 500 short prompts, addressing the needs of both novice users who prefer brief descriptions and professional creators who require detailed specifications, and provides bilingual Chinese and English versions to systematically probe robustness to both prompt length and language. We independently evaluate all 18 models on both language versions to verify the cross-lingual robustness of our findings.

For the evaluation pipeline, we provide \textbf{Q-Judger}, a \textbf{unified Judge Model} that scores every prompt-image pair at the finest granularity, providing attributable evaluation while avoiding the systematic biases of MLLM judges and the cost of manual review. Existing MLLM-based evaluators typically output only a single overall score, which tells users \textit{whether} an image is good but not \textit{why} it succeeds or fails on specific dimensions. In contrast, our Judge Model produces a complete fine-grained score vector across all 56 third-level facets for each sample, enabling precise diagnosis of capability gaps. The model is built on Qwen3.6-27B and trained on data densely annotated by 80 professional experts from art academies—specializing in photography, directing, and fine arts—under strict blind labeling with at least three independent reviews per sample. Scores are aggregated bottom-up from L3 to L2, from L2 to L1, and finally to an overall score, as detailed in Section \ref{sec:3.4}. The unified Judge Model achieves strong ranking consistency with human expert judgments (Spearman $\rho = 0.92$), substantially reducing evaluation cost while remaining closely aligned with professional standards. We release both the Judge Model and the complete prompt set to support fully offline T2I evaluation.

Finally, we conduct a comprehensive evaluation of 18 widely used T2I models, including GPT Image 2, Nano Banana Pro \citep{nanobananapro2025}, GPT Image 1.5 \citep{openai2025chatgptimages}, Seedream 5.0 \citep{seedream50lite}, Imagen 4.0 Ultra \citep{deepmind2025imagen}, Qwen Image 2.0 Pro~\citep{qwenimage20}, and HunyuanImage 3.0 \citep{cao2026hunyuanimage30technicalreport}, under both Chinese and English prompts, with overall results summarized in Tab.~\ref{tab:overall_performance}. Our findings reveal that GPT Image 2 achieves the highest overall score (ZH: 64.7, EN: 65.2) across all five pillars, while Qwen Image 2.0 Pro ranks fifth under both languages. Crucially, the two application-driven pillars, \textit{Creative Generation} and \textit{Real-world Fidelity}, exhibit the largest inter-model variance, confirming that they target precisely the capability gaps where existing benchmarks provide little insight. Four L3 facets (Physical Logic, Anatomical Fidelity, Animals, and Contact Interaction) emerge as systemic ceilings of current T2I models, where even the best models score below 44. Detailed English-prompt results and cross-language comparisons are provided in Appendix~\ref{sec:en_results}.

%% file: content/section_related_work.tex
\textbf{T2I Benchmarks.} Early benchmarks such as GenEval \citep{ghosh2023genevalobjectfocusedframeworkevaluating} and T2I-CompBench \citep{huang2025t2icompbenchenhancedcomprehensivebenchmark} focus on evaluating basic capabilities like object-centric properties, typically leveraging detectors and similarity models for scalable evaluation. Recent benchmarks adopt VQA and stronger judge models to move toward verbs and richer scene semantics, like TIFA \citep{hu2023tifaaccurateinterpretabletexttoimage} and DSG-1k \citep{cho2024davidsonianscenegraphimproving}. Newer benchmarks further target instruction-following under prompt complexity (TIIF-Bench \citep{wei2025tiifbenchdoest2imodel}), prompt reliability (Gecko \citep{wiles2025revisitingtexttoimageevaluationgecko}), and world knowledge/reasoning (WISE \citep{niu2025wiseworldknowledgeinformedsemantic}, R2I-Bench \citep{chen2025r2ibenchbenchmarkingreasoningdriventexttoimage}, T2I-ReasonBench \citep{sun2025t2ireasonbenchbenchmarkingreasoninginformedtexttoimage}). Comprehensive frameworks such as UniGenBench++ \citep{wang2025unigenbenchunifiedsemanticevaluation} and OnelG-Bench \citep{chang2025oneigbenchomnidimensionalnuancedevaluation} also pursue multi-dimensional coverage, including text rendering and stylization. Despite their breadth, these benchmarks have not yet achieved sufficient discriminative robustness for frontier models, and provide incomplete coverage of \textit{Real-world Fidelity} and \textit{Creative Generation}, the two capabilities that dominate practical T2I usage yet remain largely overlooked.

\textbf{T2I Evaluation Methods.} Early automated evaluation relied on specialized discriminative models as judges, as in GenEval \citep{ghosh2023genevalobjectfocusedframeworkevaluating}. The field has since largely converged on VQA/VLM-mediated scoring, where prompts are converted into questions and correctness is assessed by VQA consistency: TIFA \citep{hu2023tifaaccurateinterpretabletexttoimage} uses LLM-driven question generation, DSG \citep{cho2024davidsonianscenegraphimproving} enforces structured coverage and consistency constraints, and ConceptMix \citep{wu2024conceptmixcompositionalimagegeneration} grades concept presence via VLM-based questioning. In parallel, preference-based reward models (e.g., ImageReward \citep{xu2023imagerewardlearningevaluatinghuman}) and large-scale human preference datasets (e.g., Pick-a-Pic \citep{kirstain2023pickapicopendatasetuser}) aim to capture subjective quality that resists discrete correctness criteria.

However, assessing creativity and aesthetics remains fundamentally challenging due to their inherently subjective and multi-faceted nature. Existing approaches often rely on aesthetic datasets annotated with aggregate human ratings (e.g., AVA or aesthetic subsets in Pick-a-Pic), but these are typically limited in scale, cultural diversity, and artistic scope, hindering generalization across styles, domains, and creative intents. More critically, most benchmarks do not adequately disentangle technical quality (such as sharpness or artifact absence) from artistic creativity, which encompasses originality, compositional balance, color harmony, narrative coherence, and emotional impact. Professional artistic evaluation treats these aspects as interdependent yet distinct; however, current automated methods conflate them into a single score or proxy, producing ambiguous signals that poorly reflect expert judgment. Furthermore, none of these approaches are rooted in the actual creative workflow: their evaluation dimensions are not grounded in the professional standards that real creators apply when producing and reviewing images.

As a result, even advanced evaluation pipelines dominated by single or homogeneous judges remain sensitive to systematic bias and benchmark drift, and still cannot adequately assess both knowledge-consistent realism and authentic creative expression, the twin pillars of professional T2I applications.

This gap between benchmark objectives and the demands of real-world creators calls for a fine-grained, application-aligned assessment framework grounded in professional practice. To address this, we introduce \textbf{Qwen-Image-Bench}, a comprehensive benchmark designed from a creator-centric perspective. Through an expert-co-designed taxonomy, it jointly and explicitly evaluates a T2I model's capacity for \textit{Real-world Fidelity} and \textit{Creative Generation}, supported by a unified Judge Model that delivers rubric-grounded fine-grained scoring at the third-level facet granularity.

%% file: content/section_benchmark.tex
Qwen-Image-Bench is built on three tightly coupled components: a hierarchical capability taxonomy that redefines what is measured (Sec.\ref{sec:3.1}), an expert-in-the-loop prompt factory that turns the taxonomy into executable evaluation prompts (Sec.\ref{sec:3.2}), and a unified Judge Model supervised by professional annotators that scores every generated image at the finest third-level granularity and propagates those scores up the taxonomy to produce interpretable diagnostics at any level (Sec.~\ref{sec:3.3}--\ref{sec:3.4}). We describe each in turn, emphasizing the design choices that distinguish our framework from prior T2I benchmarks.

\subsection{A Hierarchical Capability Taxonomy}
\label{sec:3.1}
\textbf{Motivation.} Existing T2I benchmarks collapse generation ability into a handful of coarse axes, typically image quality, aesthetics, and text-image alignment. As frontier models converge on these fundamentals, flat taxonomies lose discriminative power and, more importantly, lose \emph{explanatory} power: they can rank models but cannot localize where they differ. What is needed is a capability structure that mirrors how professional creators reason about image generation, from the overall purpose of an image, through the sub-capabilities it draws on, down to the concrete visual details a reviewer inspects. Our taxonomy is designed from the creator's perspective (creator-centric): rather than asking \emph{what a researcher can measure}, we ask \emph{what a professional creator actually cares about in their workflow}.

\textbf{Three-level structure.} Qwen-Image-Bench organizes evaluation as a top-down capability tree with three levels of granularity. At the root sit five \textbf{first-level pillars}: the three conventional dimensions (\textit{Quality}, \textit{Aesthetics}, and \textit{Alignment}) plus two application-driven pillars, \textit{Real-world Fidelity} and \textit{Creative Generation}, that target the capabilities most valued in modern creator workflows. Each pillar is then expanded by professional artists into a set of \textbf{second-level sub-capabilities} reflecting the concrete decisions an artist makes during creation, such as World Knowledge and Fairness under Real-world Fidelity, or Visual Storytelling and Design Applications under Creative Generation. Every sub-capability is further decomposed into \textbf{third-level facets}, each defined by a precise rubric that specifies the observable criteria for scoring, making evaluation transparent and reproducible. In total, the taxonomy comprises 5 first-level pillars, 23 second-level sub-capabilities, and 56 third-level evaluation facets (Fig.~\ref{fig:teaser}; Tab.~\ref{tab:Evaluation Dimensions}).

\textbf{Why the hierarchy matters.} This design yields three useful properties. 1) Every facet is \emph{atomic and unambiguous}, so prompts, annotations, and scores share the same vocabulary. 2) Facets \emph{propagate deterministically upward}: each third-level facet belongs to exactly one second-level sub-capability and one first-level pillar, allowing fine-grained scores at all three levels from a single set of images. 3) The hierarchy is also \emph{diagnostic by construction}: a model's strength or weakness can always be traced to an identifiable artistic sub-skill rather than an opaque aggregate.

\begin{figure*}[t]
  \centering
  \includegraphics[width=\textwidth]{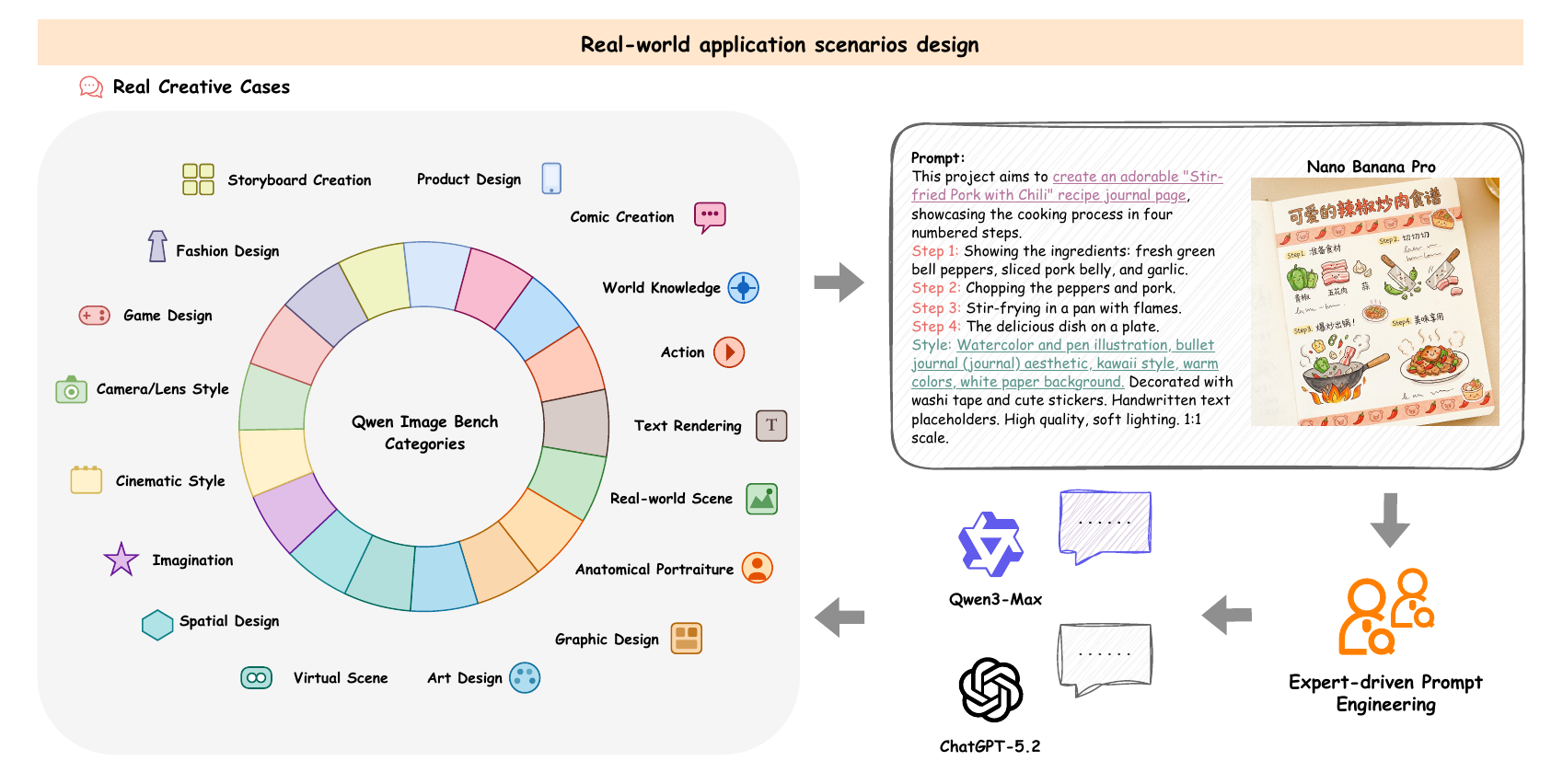}
  \caption{Pipeline for constructing real-world application prompts}
  \label{fig:pipeline}
\end{figure*}

\subsection{Expert-in-the-Loop Prompt Construction}
\label{sec:3.2}
The taxonomy provides what to evaluate; the prompt set determines whether the evaluation is actually realized. For the two application-driven pillars, Qwen-Image-Bench adopts an \textbf{expert-in-the-loop prompt factory} (Fig.~\ref{fig:pipeline}) in which professional artists and designers are not post-hoc reviewers but co-authors. Drawing on real-world creative cases, they define a prompt scope spanning \textbf{17 application domains}. Every prompt in the benchmark is thus anchored to a concrete creative scenario that real users pay for.

Formally, the factory is a four-stage pipeline tightly coupled to the three-level taxonomy:

\textbf{(1) Facet-targeted sampling.} Let $C$ denote the set of 56 third-level facets. For each prompt, we sample a target subset
\begin{equation}
  C_p = \{c_1, \dots, c_k\} \subset C, \quad k \geq 3,
\end{equation}
under a balancing constraint that each facet and each first-level pillar are visited with roughly equal frequency across the final 1{,}000 prompts. Whenever possible, $C_p$ spans at least three first-level pillars, so that a single prompt jointly stresses, for example, alignment accuracy, compositional aesthetics, and real-world fidelity.

\textbf{(2) Bilingual drafting.} Given $C_p$, we prompt large language models (Qwen3-Max and ChatGPT-5.2) to generate semantically aligned English and Chinese drafts together with explicit instantiation notes:
\begin{equation}
  (p_{\mathrm{en}}, p_{\mathrm{zh}}, \{d_i\}_{i=1}^{k}) \sim \mathrm{LLM}_{\mathrm{gen}}(C_p),
\end{equation}
where $d_i$ records precisely how facet $c_i$ is realized in the drafts $p_{\mathrm{en}}$ and $p_{\mathrm{zh}}$. This forces the mapping between prompt content and evaluation targets to be explicit rather than inferred.

\textbf{(3) Expert review, rewrite, and re-verification.} Each draft passes through a professional gate in which artists may: discard prompts that are trivial, ambiguous, or insufficiently discriminative; rewrite prompts whose instantiation of $C_p$ is weak or generic; and re-verify $\{(c_i, d_i)\}$ against the rubric so that every targeted facet is genuinely exercised. Only prompts that survive this gate enter the benchmark, concretely grounding difficulty and realism in authentic creative standards.

\textbf{(4) Length-variant expansion.} To probe robustness to linguistic complexity, prompts are split into short and long variants using thresholds of 70 characters (Chinese) and 235 characters (English). Long variants are produced by LLM expansion under a semantics-preserving constraint $r$,
\begin{equation}
  \tilde{p} \sim \mathrm{LLM}_{\mathrm{expand}}(p \mid r),
\end{equation}
after which the facet mapping is re-aligned so that added scene and attribute detail remains rubric-consistent:
\begin{equation}
  \{(\hat{c}_j, \hat{d}_j)\}_{j=1}^{k'} \sim \mathrm{LLM}_{\mathrm{align}}\bigl(\tilde{p}, \{(c_i, d_i)\}_{i=1}^{k}\bigr), \quad k' \leq k + 5.
\end{equation}

\textbf{Final prompt set.} This pipeline yields 1{,}000 prompts, evenly split into 500 short and 500 long, each released in Chinese and English, enabling controlled comparison across language, length, and capability composition. On average, each prompt jointly exercises facets spanning 3 to 5 first-level pillars, ensuring that multiple capability dimensions are tested in a single inference pass.

\subsection{Unified Judge Model with Rubric-Grounded Fine-Grained Scoring}
\label{sec:3.3}
\textbf{Limitations of existing evaluation paradigms.} A carefully designed taxonomy is only as useful as the scorer that respects it, and here existing automated evaluators fall short in two complementary ways. First, most of them compress quality, alignment, aesthetics, and higher-level creative capabilities into a single general-purpose judgment, which lets dominant signals such as aesthetics or semantic alignment silently drown out subtler capabilities like world knowledge or creative intent. Second, and more critically, a growing share of recent evaluation pipelines skips human supervision altogether and trains reward models directly on labels produced by a single MLLM judge. UniGenBench++~\cite{wang2025unigenbenchunifiedsemanticevaluation}, for instance, adopts Gemini-2.5 as both its evaluator and its label source for the reward model, so whatever systematic biases, blind spots, or aesthetic preferences Gemini-2.5 carries are directly transcribed into the evaluator and, by extension, into every downstream score. This pipeline yields neither an objective standard nor a professionally grounded one; it merely reproduces one MLLM's worldview at scale.

\textbf{Unified Judge Model with human supervision.} Qwen-Image-Bench departs from both practices at once: we commit to a \emph{single unified Judge Model supervised by large-scale, expert human annotation at the finest granularity}, with all training labels produced by professional aesthetic and technical annotators under a rubric-driven protocol rather than by any model. Our \textbf{Q-Judger} scores every prompt-image pair across all 56 third-level facets, so each dimension receives an independent, rubric-grounded judgment rather than being collapsed into an opaque aggregate. This fine-grained design makes every capability shortcoming explicit, enables precise cross-model comparison at the sub-skill level, and directly informs targeted data augmentation for future model improvement.

\textbf{Scoring protocol.} We introduce a single \textbf{Judge Model} (Q-Judger) $J$ fine-tuned from the multimodal foundation model Qwen3.6-27B. Given a prompt-image pair $(p, I)$, $J$ consumes the image together with the prompt and the complete taxonomy-aware checklist, and outputs a structured score for \emph{every} third-level facet $c \in C$:
\begin{equation}
  \{s_c(p, I)\}_{c \in C} = J(p, I, \{r_c\}_{c \in C}),
\end{equation}
where $r_c$ denotes the rubric for facet $c$. Each score $s_c$ takes one of four values:
\begin{itemize}[leftmargin=1.5em, topsep=2pt, itemsep=1pt]
  \item \textbf{0 (Fail)}: A clear defect is present that would noticeably reduce image quality.
  \item \textbf{1 (Pass)}: No defect is observable; the image meets baseline expectations.
  \item \textbf{2 (Excel)}: Exceptionally executed; concrete excellence is observable.
  \item \textbf{N/A}: This criterion does not apply to the given image or prompt.
\end{itemize}
By scoring every facet independently, the Judge Model ensures that no single dominant signal drowns out subtler dimensions, and that each capability receives a dedicated, rubric-verifiable assessment.

\textbf{Rubric-grounded annotation with taxonomy-aware context.} We recruit a pool of \textbf{80 professional annotators}, all holding degrees from art academies with backgrounds in photography, directing, or fine arts. The annotation process is deliberately structured to be checklist-driven and taxonomy-aware. All model identities are stripped before annotation (blind labeling), and the 80 annotators are randomly assigned to different prompt-image pairs so that each pair is independently scored by at least \textbf{three experts} who cross-review one another's judgments. For every training pair $(p, I)$, the prompt's pre-assigned third-level facets (together with the second-level sub-capabilities they roll up into) and their concrete rubrics are exposed to the annotator for scoring. This ``facet-by-facet, rubric-by-rubric'' protocol guarantees that every training label is anchored to concrete, verifiable evidence rather than gestalt impression.
\textbf{Annotation scale.} The Judge Model is trained on over \textbf{130{,}000} bilingual human-labeled prompt-image pairs, sampled to balance model sources, difficulty levels, and third-level facet coverage. In aggregate, the training set provides densely supervised signals across all five pillars, with each fine-grained facet receiving substantial supervision to support reliable fine-grained prediction.

\subsection{Evaluation Pipeline and Multi-granularity Scoring}
\label{sec:3.4}
\textbf{Fine-grained scoring at the third level.} At inference, the Judge Model evaluates every prompt-image pair $(p, I)$ across the complete set of 56 third-level facets. For each facet $c$, it produces a score $s_c(p, I) \in \{0, 1, 2, \text{N/A}\}$ according to the rubric-grounded protocol described in Sec.\ref{sec:3.3}. Facets that are genuinely inapplicable to a given prompt (e.g., ``Fashion Styling'' for a landscape prompt) receive N/A and are excluded from downstream aggregation. Formally,
\begin{equation}
  s_c(p, I) = J_c(p, I, r_c),
\end{equation}
where $J_c$ denotes the Judge Model's output for facet $c$ under rubric $r_c$.

\textbf{Score normalization.} Before aggregation, every non-N/A third-level score is mapped to a unified $[0, 100]$ scale:
\begin{equation}
  \phi(s) = \begin{cases} 0 & \text{if } s = 0 \\ 60 & \text{if } s = 1 \\ 100 & \text{if } s = 2 \\ \text{N/A} & \text{otherwise} \end{cases}
\end{equation}
The mapping is deliberately non-linear: 1 (Pass) maps to 60 rather than the midpoint 50, placing it at the conventional ``passing threshold'' in a percentage-graded system. This design amplifies the gap between Fail and Pass (0 vs.\ 60) relative to the gap between Pass and Excel (60 vs.\ 100), reflecting the practical reality that the distinction between unacceptable and acceptable output is far more consequential for user experience than the distinction between acceptable and excellent.

\textbf{Multi-granularity aggregation.} A distinctive property of our pipeline is that the taxonomy is preserved throughout scoring, so a \emph{single} pass of Judge Model inference produces scores at every level of the hierarchy without any additional annotation or re-inference. The aggregation proceeds bottom-up:

\textbf{Per-sample hierarchical aggregation.} For each prompt-image pair $(p, I)$, scores are aggregated bottom-up through the taxonomy. Let $\mathcal{C}(p)$ denote the set of L3 facets activated by prompt $p$. The sample-level L2 score for sub-capability $b$ is the mean of its active child facets:
\begin{equation}
  s_b(p, I) = \frac{1}{|\{c \in b : c \in \mathcal{C}(p)\}|} \sum_{\substack{c \in b \\ c \in \mathcal{C}(p)}} \phi(s_c(p, I)).
\end{equation}
The sample-level L1 score for pillar $d$ averages its active child sub-capabilities:
\begin{equation}
  s_d(p, I) = \frac{1}{|\mathcal{B}_d(p)|} \sum_{b \in \mathcal{B}_d(p)} s_b(p, I),
\end{equation}
where $\mathcal{B}_d(p) = \{b \in \mathcal{B}_d : b \text{ has at least one active facet in } \mathcal{C}(p)\}$. The sample-level overall score is the unweighted mean of the active pillar scores:
\begin{equation}
  s_{\text{overall}}(p, I) = \frac{1}{|\mathcal{D}(p)|} \sum_{d \in \mathcal{D}(p)} s_d(p, I),
\end{equation}
where $\mathcal{D}(p) \subseteq \mathcal{D}$ is the set of pillars activated by prompt $p$ (ranging from 3 to 5).

\textbf{Model-level scores.} The benchmark-level score at each granularity is obtained by averaging the corresponding sample-level scores over all prompts that activate that component:
\begin{equation}
  S_c = \frac{1}{|\mathcal{T}_c|} \sum_{(p,I) \in \mathcal{T}_c} \!\phi(s_c(p,I)), \quad
  S_b = \frac{1}{|\mathcal{T}_b|} \sum_{(p,I) \in \mathcal{T}_b} \!s_b(p,I), \quad
  S_d = \frac{1}{|\mathcal{T}_d|} \sum_{(p,I) \in \mathcal{T}_d} \!s_d(p,I),
\end{equation}
where $\mathcal{T}_c$, $\mathcal{T}_b$, $\mathcal{T}_d$ denote the sets of prompt-image pairs that activate facet $c$, sub-capability $b$, and pillar $d$, respectively. The overall model score averages across all prompts:
\begin{equation}
  S_{\text{overall}} = \frac{1}{|\mathcal{P}|} \sum_{(p,I) \in \mathcal{P}} s_{\text{overall}}(p,I).
\end{equation}
All scores naturally reside in $[0, 100]$.

\begin{algorithm}[t]
\caption{Qwen-Image-Bench Multi-Granularity Scoring}
\label{alg:scoring}
\begin{algorithmic}[1]
\Require Prompt set $\mathcal{P}$, Image set $\mathcal{I}$ (generated by model $m$), Taxonomy $\mathcal{T}$, Judge Model $J$
\Ensure $S_{\text{overall}}$, $\{S_d\}$, $\{S_b\}$, $\{S_c\}$
\For{each prompt-image pair $(p, I) \in \mathcal{P} \times \mathcal{I}$}
    \State Construct taxonomy-aware checklist $\mathcal{C}(p)$ from $\mathcal{T}$
    \State $\{s_c(p,I)\}_{c \in \mathcal{C}(p)} \leftarrow J(p, I, \mathcal{C}(p))$ \Comment{Judge Model inference}
    \For{each facet $c \in \mathcal{C}(p)$}
        \State $\hat{s}_c(p,I) \leftarrow \phi(s_c(p,I))$ \Comment{Map to $[0,100]$}
    \EndFor
\EndFor
\State \textbf{// Per-sample bottom-up aggregation}
\For{each prompt-image pair $(p, I)$}
    \For{each active L2 sub-capability $b \in \mathcal{B}(p)$}
        \State $s_b(p,I) \leftarrow \text{mean}(\{\hat{s}_c(p,I) : c \in \text{children}(b) \cap \mathcal{C}(p)\})$
    \EndFor
    \For{each active L1 pillar $d \in \mathcal{D}(p)$}
        \State $s_d(p,I) \leftarrow \text{mean}(\{s_b(p,I) : b \in \text{children}(d) \cap \mathcal{B}(p)\})$
    \EndFor
    \State $s_{\text{overall}}(p,I) \leftarrow \text{mean}(\{s_d(p,I) : d \in \mathcal{D}(p)\})$ \Comment{Mean over active pillars}
\EndFor
\State \textbf{// Model-level scores (mean across prompts)}
\State $S_c \leftarrow \text{mean}(\{\hat{s}_c(p,I) : (p,I) \in \mathcal{T}_c\})$ for each L3 facet $c$
\State $S_b, S_d, S_{\text{overall}} \leftarrow$ analogous prompt-level means
\State \Return $S_{\text{overall}}$, $\{S_d\}$, $\{S_b\}$, $\{S_c\}$
\end{algorithmic}
\end{algorithm}

which naturally falls in $[0, 100]$ and provides a single headline metric for model comparison. This bottom-up design ensures that every aggregate score is \emph{numerically consistent} with its constituent fine-grained scores: a model's first-level pillar score can always be traced to the specific third-level facets where it excels or falls short, and its overall ranking can be decomposed into actionable sub-capability and facet-level diagnostics.

%% file: content/section_experiment.tex
\subsection{Implementation Details of the Judge Model}

\textbf{Backbone and framework.} Our \textbf{Q-Judger} is initialized from Qwen3.6-27B and fine-tuned within the MS-SWIFT \cite{zheng2024llamafactory} framework, which we adopt for its mature multimodal fine-tuning support and reproducibility. Q-Judger can be found in \url{https://huggingface.co/Qwen/Qwen-Image-Bench}.

\textbf{Inference.} At evaluation time, we use deterministic decoding with \texttt{seed}$\,{=}\,42$, \texttt{repetition\_penalty}$\,{=}\,1.05$, \texttt{top\_k}$\,{=}\,1$, \texttt{top\_p}$\,{=}\,1$, and \texttt{temperature}$\,{=}\,0$. We enable the model's thinking mode (\texttt{enable\_thinking}$\,{=}\,\text{True}$) to allow chain-of-thought reasoning before scoring. The Judge Model receives a structured prompt consisting of a system-level role definition, the generation prompt, the generated image, a scoring rubric, and a dimension-specific evaluation checklist encoding the full L2/L3 hierarchy. The complete prompt templates and all five pillar-level checklists are provided in Appendix~\ref{sec:appendix_prompts}.

\textbf{Training data.} \textbf{Q-Judger} is trained on over \textbf{130{,}000} bilingual expert-annotated prompt-image pairs following the rubric-grounded protocol described in Sec.\ref{sec:3.3} (80 art-academy annotators, blind labeling, at least three independent reviews per sample).

\input{tables/table_spearman}

\textbf{Alignment with human experts.} On a held-out evaluation set, senior expert reviewers independently rate each image on a 1--10 holistic scale per L1 pillar; models are then ranked by their per-pillar and overall average human scores (detailed results shown in Table \ref{tab:human_ranking}). The Judge Model achieves strong ranking consistency with these human rankings (as shown in Table \ref{tab:spearman}), confirming that the Judge Model's fine-grained scores faithfully track professional judgment across all evaluation dimensions.

\subsection{Benchmarked T2I Models}

We evaluate 18 representative T2I models on Qwen-Image-Bench, spanning a broad spectrum of frontier T2I models: GPT Image 2, GPT Image 1.5~\cite{openai2025chatgptimages}, GPT Image 1~\cite{openai2024gpt4ocard}, Nano Banana 2.0~\cite{nanobanana2}, Nano Banana Pro~\cite{nanobananapro2025}, Seedream 5.0~\cite{seedream50lite}, Seedream 4.5~\cite{bytedance2026seedream45}, Seedream 4.0~\cite{seedream2025seedream40nextgenerationmultimodal}, Imagen 4.0 Ultra~\cite{deepmind2025imagen}, Imagen 4.0~\cite{deepmind2025imagen}, FLUX 2 Max~\cite{blackforestlabs2025flux2max}, FLUX 2 Pro~\cite{blackforestlabs2025flux2}, Qwen Image 2.0 Pro~\cite{qwenimage20}, Qwen Image 2512~\cite{wu2025qwenimagetechnicalreport}, Qwen Image~\cite{wu2025qwenimagetechnicalreport}, HunyuanImage 3.0~\cite{cao2026hunyuanimage30technicalreport}, GLM Image~\cite{zai2026glmimage}, and Kling Image 2.1~\cite{klingai2026imagestylize}.

For every model, we generate images for all 1{,}000 Qwen-Image-Bench prompts, then evaluate all prompt-image pairs with the Judge Model, which produces a structured score vector of 56 third-level facet scores per pair. These fine-grained scores are aggregated bottom-up following the protocol in Sec.\ref{sec:3.4} to yield first-level pillar scores and the overall benchmark score.

\input{tables/table_qwen_image_bench}

\subsection{Overall Performance on Qwen-Image-Bench}
\label{sec:overall}

Tab.~\ref{tab:overall_performance} reports the performance of all 18 T2I models along the five first-level pillars and the aggregated overall score. We present the results following a top-down analysis: overall ranking, fine-grained capability profiles, variance-driven discriminability analysis, sub-capability rankings, per-pillar rankings, industry-wide capability landscape, and cross-tier gap analysis.

\begin{figure}[t]
\centering
\includegraphics[width=\linewidth]{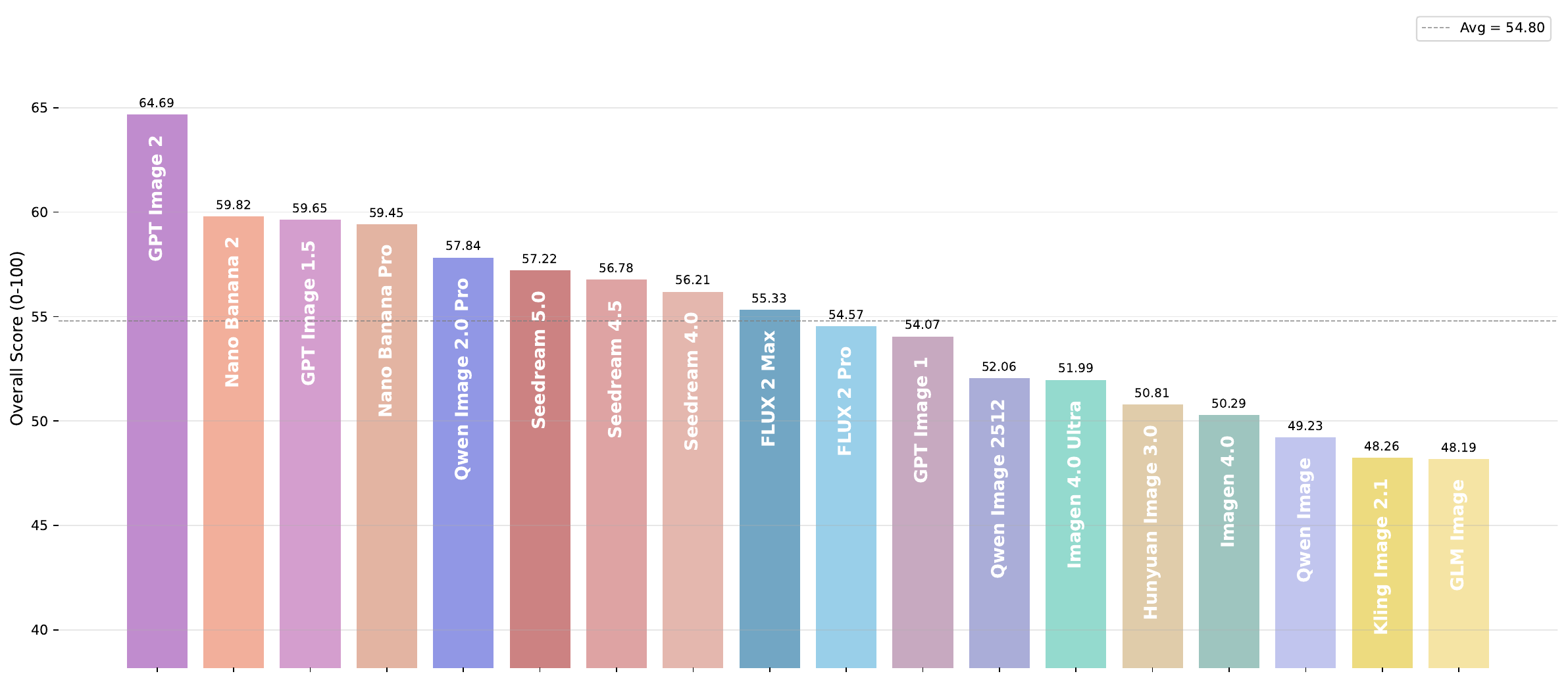}
\caption{Overall score ranking of 18 T2I models on Qwen-Image-Bench. The dashed line marks the cross-model mean. The English-prompt ranking (Appendix, Fig.~\ref{fig:overall_ranking_en}) preserves the same five-tier structure.}
\label{fig:overall_ranking}
\end{figure}

\textbf{Overall Ranking.}
Fig.~\ref{fig:overall_ranking} presents the overall ranking. GPT Image 2 leads by nearly 5 points over the second-ranked Nano Banana 2.0, with GPT Image 1.5 and Nano Banana Pro following closely to form a tightly clustered second tier. Qwen Image 2.0 Pro ranks fifth overall, heading the third tier. GLM Image sits at the bottom, yielding a 16.5-point spread from the leader and demonstrating the benchmark's effective discriminative power across the full model spectrum. The 18 models naturally separate into five tiers: T1 (64+: GPT Image 2), T2 (59--60: Nano Banana 2.0, GPT Image 1.5, Nano Banana Pro), T3 (56--58: Qwen Image 2.0 Pro, Seedream 5.0, Seedream 4.5, Seedream 4.0), T4 (54--56: FLUX 2 Max, FLUX 2 Pro, GPT Image 1), and T5 (48--52: seven remaining models). Notably, GPT Image 2 achieves the highest score on all five L1 pillars simultaneously, forming a dominant profile with no discernible weakness, a rarity among the 18 evaluated models. Under English prompts, the same five-tier structure holds; the only within-tier change is that GPT Image 1.5 and Nano Banana 2.0 swap ranks~2 and~3 (Appendix, Fig.~\ref{fig:overall_ranking_en}).

\subsection{Fine-grained Capability Profiles}

\begin{figure*}[t]
\centering
\includegraphics[width=0.8\linewidth]{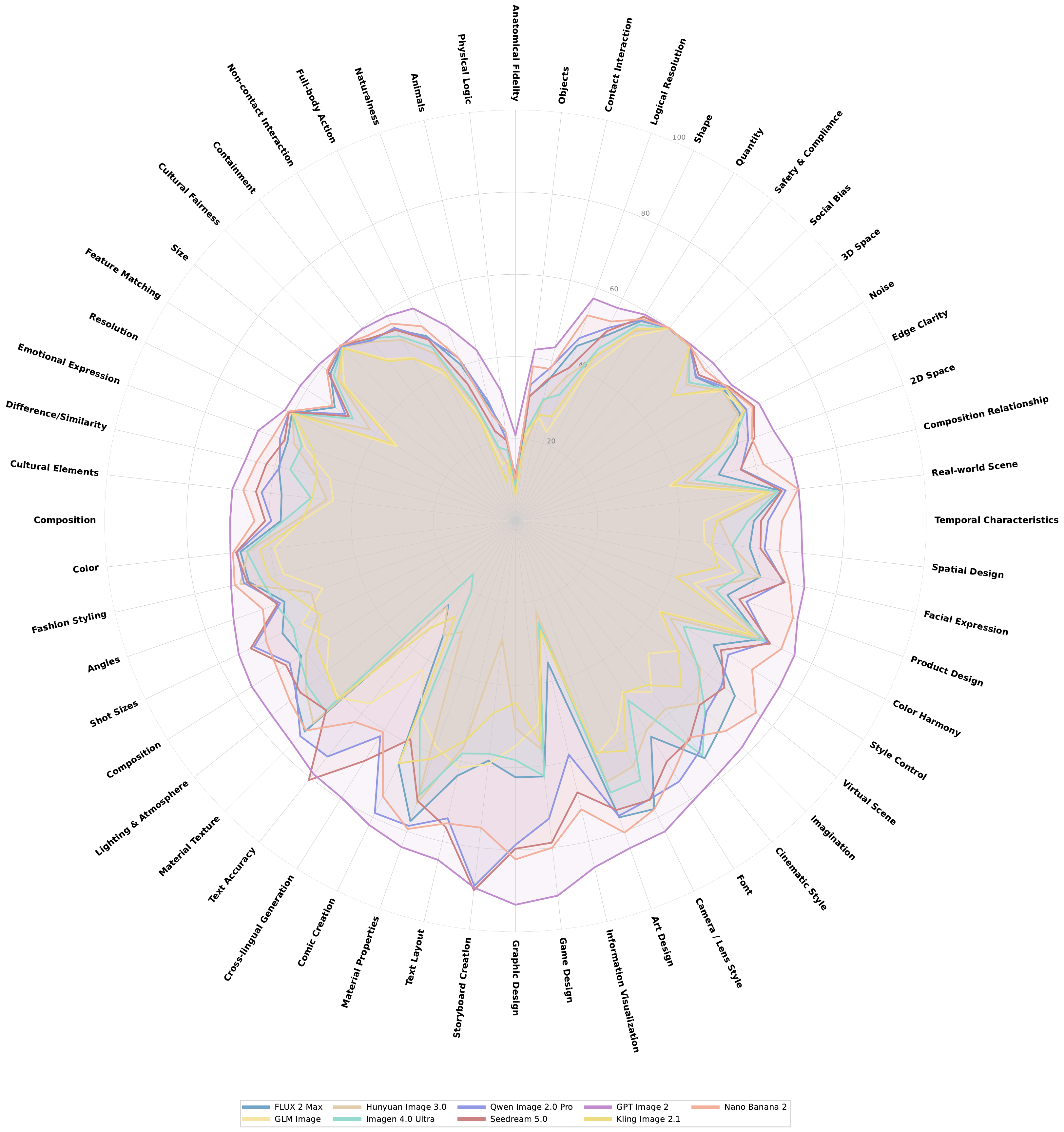}
\caption{L3-level radar chart showing model capability profiles across all 56 third-level facets, sorted by overall mean score. Each polygon represents one model; larger area indicates stronger overall capability.}
\vspace{-4mm}
\label{fig:radar_l3}
\end{figure*}

The L3 radar chart (Fig.~\ref{fig:radar_l3}) provides the most fine-grained view of model capabilities across all 56 third-level facets. We arrange the 56 facets in a \emph{heart-shaped layout} based on GPT Image 2's L3 scores: the weakest facets are placed at the heart's indentation (top), while the strongest facets converge at the heart's apex (bottom). This causes the frontier model's polygon to assume a distinctive \textbf{heart shape}: the apex encodes peak strengths, and the top-center indentation marks universal weaknesses shared across all models. Several structural patterns emerge from this visualization.

\textbf{The heart-shaped silhouette reveals capability architecture at a glance.} GPT Image 2's polygon forms the outermost heart contour, with the apex at 6-o'clock driven by five peak-scoring facets: Text Layout, Storyboard Creation, Graphic Design, Game Design, and Information Visualization (all above 84, with Graphic Design and Game Design exceeding 90). Unlike other models, its smooth and symmetric lobes flanking the apex reflect consistent superiority across both application-driven pillars rather than a few outlier strengths.

\textbf{Tier-separating facets cluster near the apex.} Three facets near the heart's tip---Text Accuracy, Cross-lingual Generation, and Information Visualization---act as primary tier discriminators. On these facets, only a handful of models (GPT Image 2, Seedream 5.0, Qwen Image 2.0 Pro, Nano Banana 2.0) sustain scores above 60, while lower-tier models collapse sharply inward (often below 35), producing a visible ``pinch'' in the heart contour just beside the apex. This concentrated inward collapse is the geometric signature of tier separation: models that maintain lateral fullness near 6-o'clock belong to the upper tiers, while those that deflate belong to the lower tiers.

\textbf{The top-center indentation exposes systemic ceilings.} Because these facets are clustered at the top by design, the shared indentation is immediately visible as an industry-wide blind spot. The heart's characteristic inward notch at the 12-o'clock position corresponds to five facets where \emph{all} models collapse simultaneously: Anatomical Fidelity (best 20.9), Physical Logic (best 31.9), Objects (best 41.9), Animals (best 42.6), and Contact Interaction (best 43.3). Notably, these span four different pillars (Aesthetics, Quality, Real-world Fidelity (Objects and Animals), and Alignment), indicating that the ceilings are not confined to a single evaluation axis but reflect a common weakness while handling fine-grained physical and biological structure or logic. 

\textbf{Mid-tier models exhibit asymmetric heart deformation.} While top-ranked models produce full, symmetric hearts, mid-tier models display progressively more irregular and deflated shapes. Qwen Image 2.0 Pro, for instance, maintains competitive lateral coverage on visual-style facets (Art Design, Camera/Lens Style, Cinematic Style, Graphic Design) but shows pronounced inward collapse on creative-precision facets located between the lobes (Text Accuracy, Font, Product Design, Comic Creation, Game Design, Information Visualization). This ``imagination vs.\ execution precision'' gap is quantitatively stark: its top-10 Creative Generation facets average 76.5 while its bottom-5 facets average only 55.5, a gap of over 21 points within the same pillar. The heart metaphor makes this immediately readable: lateral strength with medial weakness signals a model that can \emph{imagine} but struggles to \emph{execute precisely}. The same structural patterns hold under English prompts (Appendix, Fig.~\ref{fig:radar_l3_en}).

\subsection{Discriminative Power of Creator-Centric Dimensions}
\label{sec:4.5}
\begin{figure}[t]
\centering
\includegraphics[width=\linewidth]{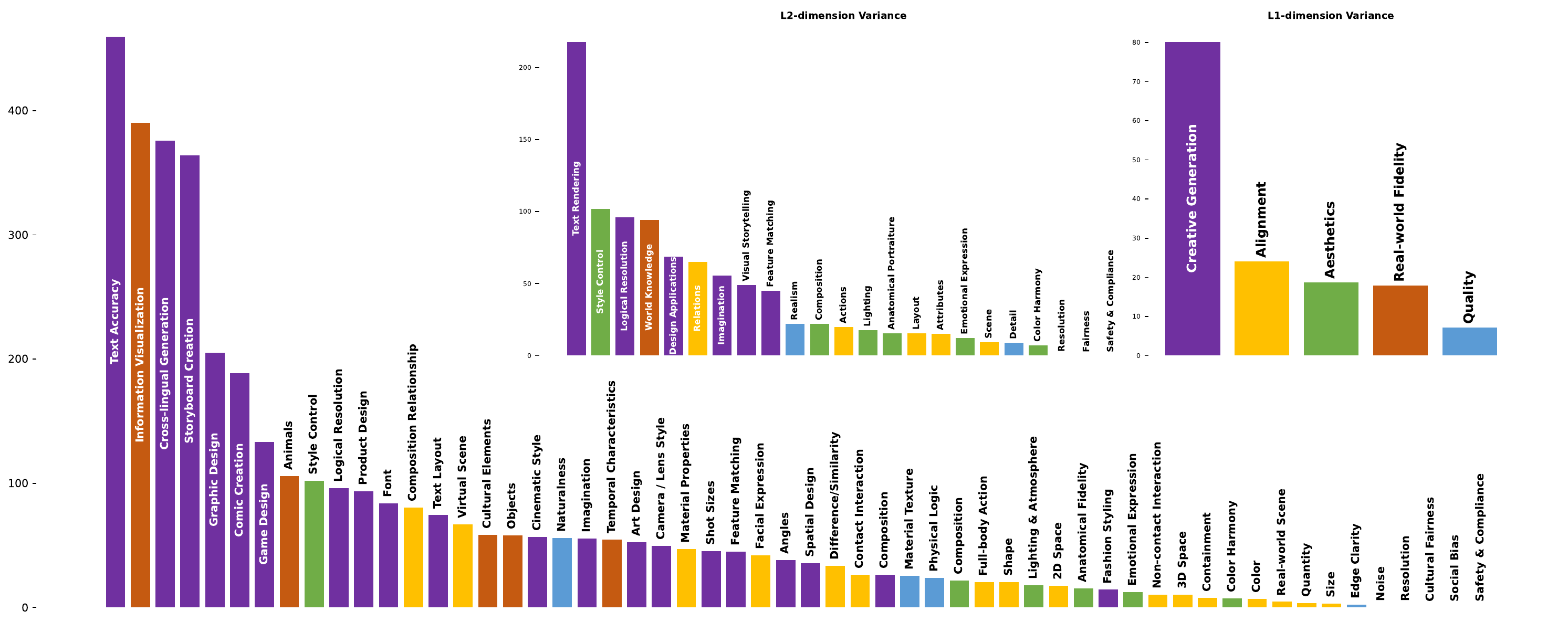}
\caption{Inter-model score variance across the three levels of the taxonomy. Main plot: L3-level variance (56 facets); insets: L2-level (23 sub-capabilities) and L1-level (5 pillars).}
\label{fig:variance}
\end{figure}

A key design goal of Qwen-Image-Bench is to introduce evaluation creator-centric dimensions that expose capability gaps invisible to previous benchmarks. Figure~\ref{fig:variance} quantifies this by measuring inter-model score variance (the higher the variance, the more a dimension differentiates models) across all three taxonomy levels.

\textbf{L3 variance pinpoints the sharpest frontiers.} At the finest granularity, Text Accuracy (under Creative Generation) is the single most discriminative facet. Information Visualization (under Real-world Fidelity) and Cross-lingual Generation (under Creative Generation) rank second and third. Of the 15 highest-variance L3 facets, 12 belong to Creative Generation or Real-world Fidelity (e.g., Storyboard Creation, Graphic Design, Cross-lingual Generation, Game Design), dimensions that jointly test creative imagination, logical reasoning, and creation execution precision.

\textbf{L2 variance confirms application-driven dimensions dominate.} Rolling up to the second level, the highest-variance L2 sub-capability is Text Rendering (under Creative Generation), followed by Style Control (under Aesthetics), Logical Resolution (under Creative Generation), and World Knowledge (under Real-world Fidelity). Among the top six L2 dimensions by variance, four belong to the two application-driven pillars introduced by our benchmark.

\textbf{L1 variance reveals where differentiation lies.} At the pillar level, Creative Generation variance is over 11$\times$ that of Quality and over 4$\times$ that of Aesthetics. The low variance on Quality indicates that basic image quality has become a ``table-stakes'' capability, while Creative Generation, the pillar most unique to our creator-centric design, is precisely where models diverge most sharply.

The variance analysis provides strong quantitative evidence that the two application-driven pillars introduced by Qwen-Image-Bench, namely Creative Generation and Real-world Fidelity, precisely target the capability gaps that existing benchmarks have not revealed. Models achieve near-consensus on basic quality and alignment (low variance), but diverge sharply on knowledge-grounded and creative tasks. These dimensions not only increase the benchmark's discriminative power but also highlight concrete directions for future model improvement. The same variance hierarchy holds under English prompts (Appendix, Fig.~\ref{fig:variance_en}), confirming that these findings are language-independent.

\subsection{Sub-capability and Facet-level Rankings}

Beyond overall and pillar-level rankings, the three-level taxonomy allows us to re-aggregate the same Judge Model outputs to produce rankings at 23 second-level sub-capabilities and 56 third-level facets without any additional inference. We highlight the creator-most-caring dimensions with the high inter-model variance, which most sharply differentiate current T2I models.
\begin{figure*}[t]
\centering
\includegraphics[width=\linewidth]{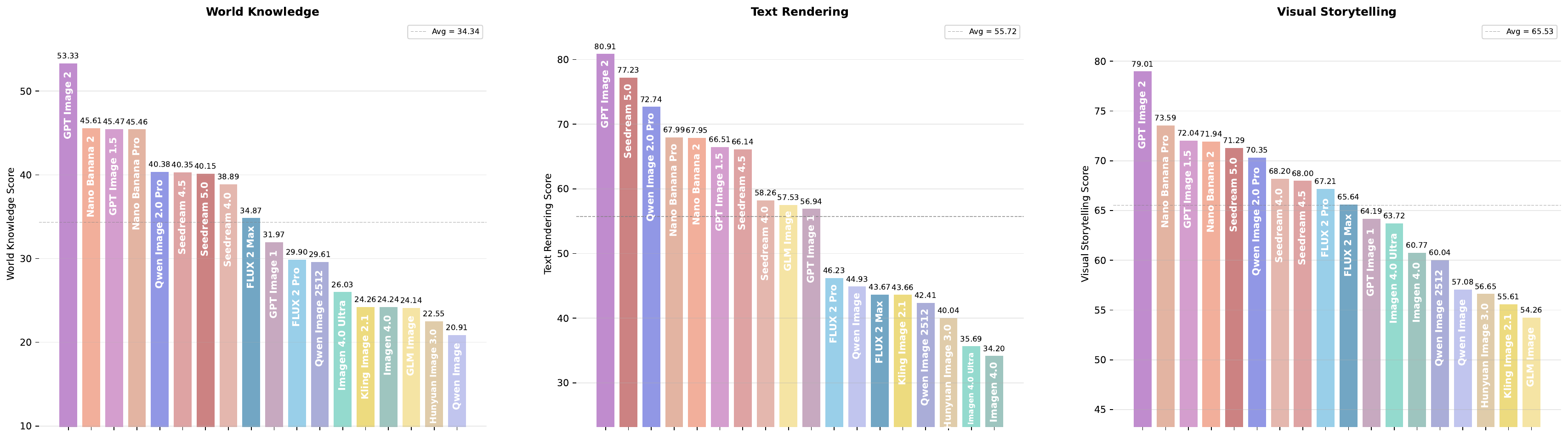}
\vspace{1mm}
\centerline{\small (a) Per-model rankings on the three highest-variance L2 sub-capabilities.}
\vspace{6mm}
\includegraphics[width=\linewidth]{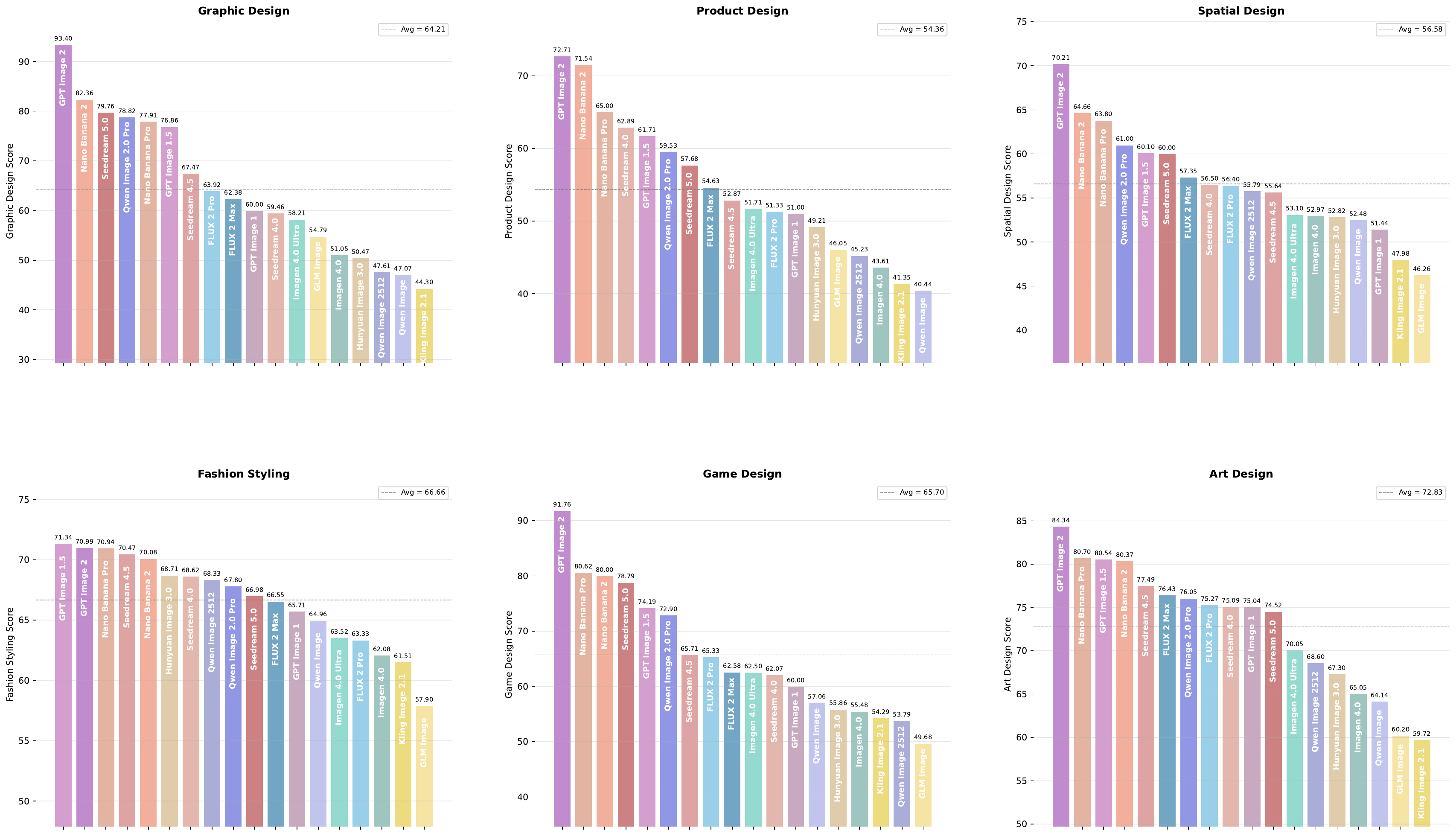}
\vspace{1mm}
\centerline{\small (b) Per-model rankings on the six L3 facets under Design Applications.}
\caption{Sub-capability and facet-level rankings. (a)~L2 rankings on the three sub-capabilities with the high inter-model variance: Text Rendering (under Creative Generation), World Knowledge (under Real-world Fidelity), and Visual Storytelling (under Creative Generation). (b)~L3 rankings on the six facets under Design Applications: Graphic Design, Product Design, Spatial Design, Fashion Styling, Game Design, and Art Design.}
\label{fig:l2l3_ranking}
\end{figure*}

\textbf{L2 rankings on the three high-variance sub-capabilities.}
Fig.~\ref{fig:l2l3_ranking}(a) shows model rankings on three L2 sub-capabilities that are both creator-most-caring and exhibit relatively high inter-model variance: Text Rendering, World Knowledge, and Visual Storytelling.

On \textit{World Knowledge} (under Real-world Fidelity), GPT Image 2 leads by an 8-point gap over the second tier. Qwen Image 2.0 Pro ranks fifth, while models below rank~8 fall sharply, with the bottom cluster scoring under half the leader's score. This dimension captures the ability to faithfully reproduce real-world objects, animals, and structured visual information, confirming it as a critical frontier for knowledge-grounded generation, remaining challenging for many models.

\textbf{L3 rankings on the six Design Applications facets.}
Fig.~\ref{fig:l2l3_ranking}(b) drills into the six L3 facets under Design Applications to reveal fine-grained specialization patterns.

\textit{Game Design} exhibits the most extreme threshold effect: GPT Image 2 leads at 91.8, followed by a tight cluster around 80 (ranks 2--4), then a sharp drop from rank~7 onward (below 66). 

\textit{Graphic Design} follows a similar pattern, with the gap from rank~1 to the median model exceeding 30 points. This ``capable or incapable'' pattern is characteristic of high-level creative tasks that require both visual imagination and domain-specific knowledge.

\textit{Fashion Styling} shows the flattest ranking gradient (a 13-point spread from leader to bottom), indicating moderate but consistent capability across models. In contrast, \textit{Product Design} produces more spread (32-point gap), with models below rank~13 scoring under 50. 

On \textit{Art Design}, ranks 2--4 are tightly compressed (separated by barely one point), suggesting that artistic style transfer is relatively mature across frontier models.

\subsection{Per-Pillar Model Rankings}

\begin{figure}[t]
\centering
\includegraphics[width=\linewidth]{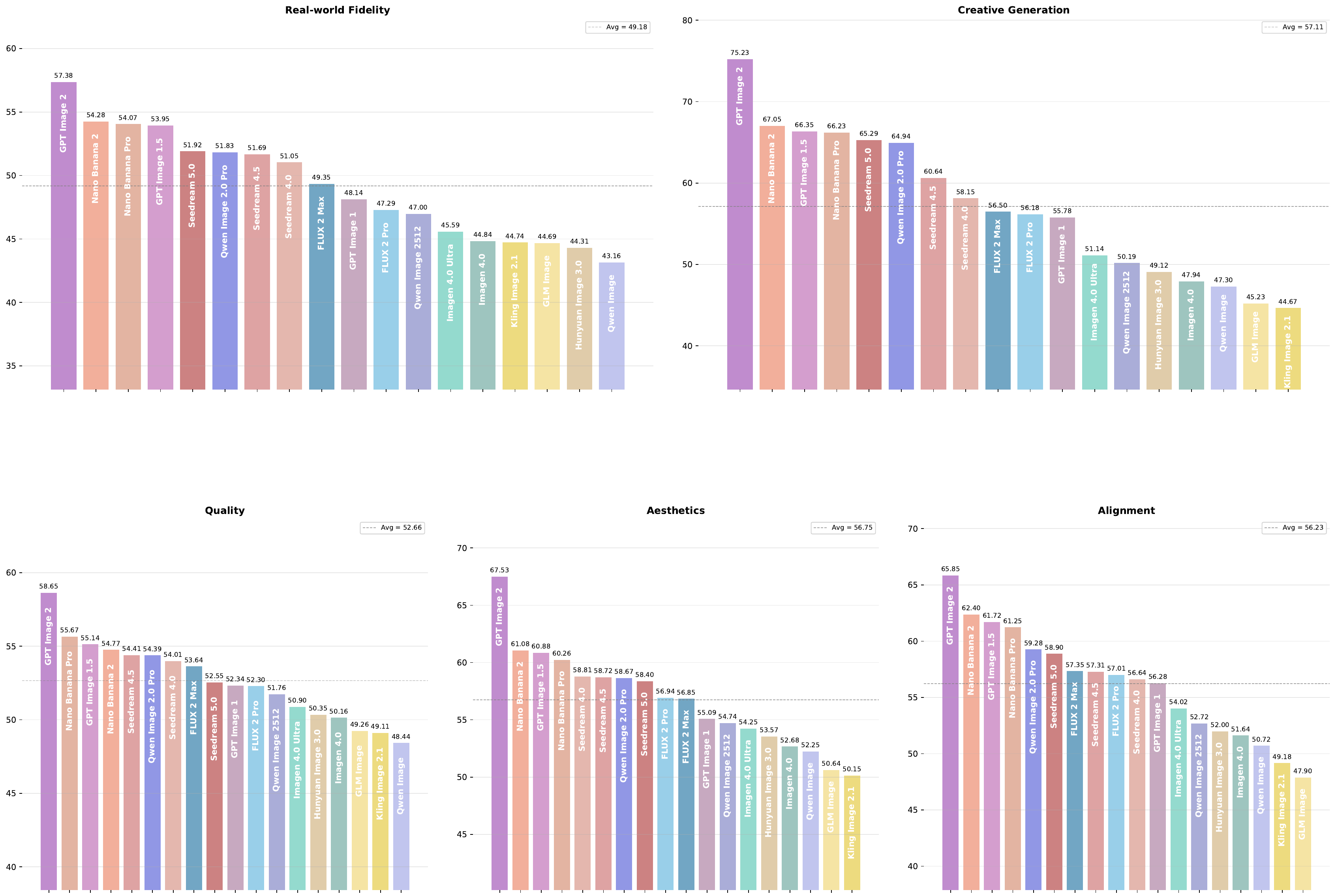}
\caption{Per-pillar model rankings across the five L1 dimensions. Model ordering shifts substantially across pillars, revealing dimension-specific strengths and weaknesses.}
\label{fig:l1_ranking}
\end{figure}

Fig.~\ref{fig:l1_ranking} presents model rankings decomposed by each of the five L1 pillars, revealing how model ordering shifts across dimensions.

\textbf{Creative Generation produces the largest ranking shifts.} GPT Image 2 leads, followed by Nano Banana 2.0 and GPT Image 1.5. Qwen Image 2.0 Pro ranks sixth on this pillar. The 30.6-point spread between the leader and the bottom-ranked model is the largest among all five pillars, confirming Creative Generation as the most discriminative dimension.

\textbf{Quality rankings diverge most from the overall leaderboard.} Nano Banana Pro climbs to second on Quality, demonstrating superior artifact suppression and physical-logic handling. The Seedream series illustrates a noteworthy version-evolution trade-off: Seedream 4.5 scores higher than 5.0 on Quality and Aesthetics, while 5.0 surges ahead on Creative Generation by over 4 points, suggesting that the newer release prioritized creative capabilities at the cost of basic image quality. This trade-off is one that our multi-pillar evaluation makes explicit but a single-score benchmark would obscure.

\textbf{Aesthetics and Alignment show the most stable rankings.} The top four models (GPT Image 2, Nano Banana 2.0, GPT Image 1.5, Nano Banana Pro) retain their positions across both pillars. On Alignment, Qwen Image 2.0 Pro rises to fifth, while on Aesthetics the Seedream series (4.0/4.5/5.0) forms a tightly clustered band with near-identical scores occupying ranks 5--8.

\textbf{Real-world Fidelity separates production-grade models.} GPT Image 2 leads, with the next cluster (Nano Banana 2.0, GPT Image 1.5, Nano Banana Pro) trailing by roughly 3 points. The 14-point gap between the leader and the lowest-performing models supports the observation that faithful reconstruction of real-world structure and knowledge-grounded content is currently a defining advantage of frontier models.

\textbf{Application-driven pillars widen the gap between tiers.} Qwen Image 2.0 Pro ranks fifth on Alignment but sixth on Quality, Real-world Fidelity, and Creative Generation (seventh on Aesthetics). Its gap to GPT Image 2 remains moderate on Quality and Alignment but widens sharply on Aesthetics and Creative Generation. Despite this, Qwen Image 2.0 Pro scores at or above the industry mean on virtually all L3 facets, indicating a solid ``no-weakness'' baseline. Per-pillar rankings under English prompts exhibit minor within-tier shifts (Appendix~\ref{sec:en_results}).

\subsection{Industry-wide Capability Landscape}

\begin{figure*}[t]
\centering
\includegraphics[width=\linewidth]{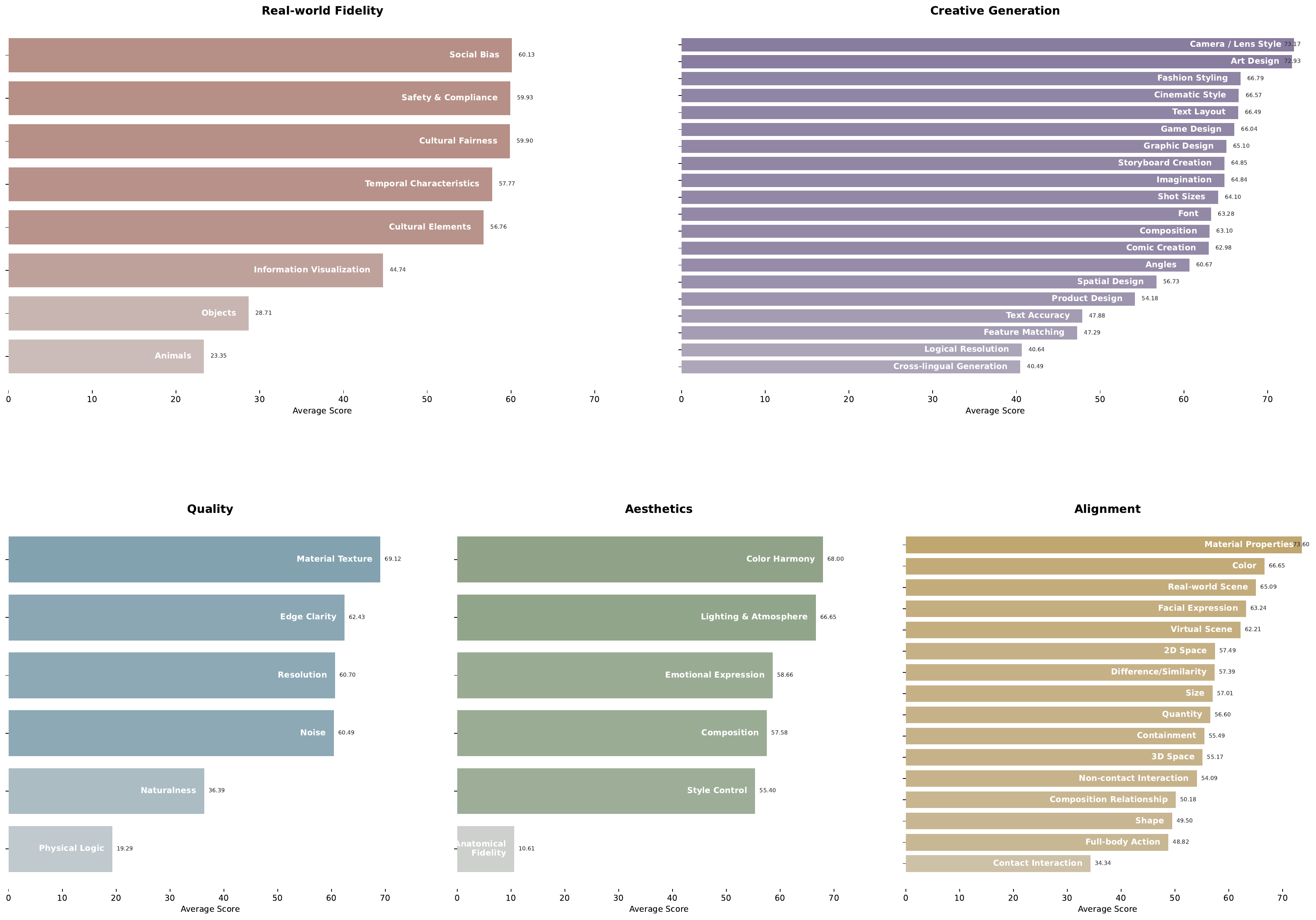}
\caption{Mean scores across all 18 models for each L3 facet, grouped by L1 pillar. Bars are sorted by descending mean within each pillar, revealing the industry-wide capability landscape for each facet.}
\label{fig:l3_bar}
\end{figure*}

Fig.~\ref{fig:l3_bar} presents the industry-wide mean score for each of the 56 L3 facets, grouped by L1 pillar and sorted by descending mean within each group. This ``capability landscape'' view reveals where the T2I field as a whole is strong, where it is adequate, and where fundamental challenges remain.

\textbf{Mature capabilities (mean $> 60$).} Facets such as Lighting \& Atmosphere, Color Harmony, Material Properties, Real-world Scene, Art Design, Camera/Lens Style, Cinematic Style, and Game Design all exceed 60, indicating that visual aesthetics, material rendering, and scene-level fidelity are relatively mature across current T2I models. Under Creative Generation, the high means mask extreme inter-model variance, meaning these are capabilities where frontier models excel but mid-tier models still struggle.

\textbf{Developing capabilities (mean 40--60).} The majority of facets fall in this range, including most Alignment dimensions (Quantity, 2D Space, Composition Relationship, Full-body Action), several Creative Generation dimensions (Text Layout, Product Design, Font, Cross-lingual Generation, Text Accuracy), and Real-world Fidelity dimensions such as Information Visualization and Objects. These facets represent the active frontier where targeted improvements can yield meaningful benchmark gains.

\textbf{Systemic ceilings (mean $< 35$).} Five facets remain industry-wide bottlenecks: Physical Logic, Anatomical Fidelity, Animals, Objects, and Contact Interaction. Even the best-performing model scores below 44 on each---the same five facets that form the 12-o'clock indentation in the radar chart (Fig.~\ref{fig:radar_l3}). Although these facets span four different L1 pillars (Quality, Aesthetics, Real-world Fidelity, and Alignment), they share a unified root cause: all require \emph{implicit world knowledge}---structural, physical, or biological---that lies beneath the visible surface of the generated image. Anatomical Fidelity and Animals, despite belonging to different pillars, both demand an internal model of biological anatomy (human or animal) that goes far beyond reproducing surface textures. Objects requires grounding in real-world three-dimensional structure. Physical Logic and Contact Interaction demand understanding of gravity, material interactions, and spatial contact relationships. In each case, the model must possess latent knowledge about \emph{how the world works}, not merely \emph{how it looks}.

This convergence reveals a structural limitation: \textbf{current models excel at reproducing visual surface patterns but remain weak at tasks requiring implicit world knowledge and logical reasoning.} Facets that reduce to learned visual statistics---Camera/Lens Style, Art Design, Cinematic Style (all above 66)---are effectively solved, because success requires only reproducing aesthetic patterns present in training data. In contrast, facets requiring implicit reasoning about structure, physics, or biology collapse uniformly, regardless of which pillar they nominally belong to. The systemic ceilings thus mark a \textbf{perception-to-cognition frontier}---models have crossed the perceptual stage (visual aesthetics, style control) but have not yet entered the cognitive stage (causal reasoning, structural understanding, commonsense grounding). Notably, even the strongest model scores below 44 on all five ceiling facets, suggesting that incremental improvements along current technical trajectories are unlikely to fundamentally break through these barriers. Closing this gap may require deeper integration of world knowledge and structured reasoning into the generation process.

\textbf{The ``perception vs.\ reasoning'' divide.} The same structural pattern manifests within Creative Generation: high-mean facets (Art Design, Camera/Lens Style, Cinematic Style) are visual-perception tasks where models leverage learned aesthetic priors, while low-mean facets (Logical Resolution, Text Accuracy, Cross-lingual Generation) require precise execution or reasoning. The over 32-point gap between the highest and lowest cross-model mean within this single pillar, combined with the largest inter-model variance among all five pillars, confirms that the ``surface reproduction vs.\ implicit reasoning'' divide is the defining structural feature of current T2I capabilities.

The L3-level and L2-level score heatmaps (Appendix, Figs.~\ref{fig:l3_heatmap}--\ref{fig:l2_heatmap}) provide complementary visualizations of these patterns at the individual-model level.

\subsection{Cross-Tier Gap Analysis and Upgrade Pathways}

The five-tier structure identified in Sec.\ref{sec:overall} raises a natural question: \emph{what specific capabilities separate adjacent tiers, and where should lower-tier models focus their improvement efforts?} Tab.~\ref{tab:tier_gap} answers this by comparing tier-averaged scores on all five L1 pillars.

\begin{table}[t]
\caption{Tier-averaged L1 pillar scores and inter-tier gaps. Application-driven pillars consistently produce larger gaps than conventional ones, confirming that they are the primary determinants of tier placement.}
\label{tab:tier_gap}

\newcolumntype{C}[1]{>{\centering\arraybackslash}p{#1}}
\centering
\begin{adjustbox}{width=0.8\linewidth}
\begin{tabular}{lC{1.6cm}C{1.6cm}C{1.6cm}C{1.6cm}C{1.6cm}|c}
\toprule
 & \multirow{2}{*}{\textit{Quality}} & \multirow{2}{*}{\textit{Aesthetics}} & \multirow{2}{*}{\textit{Alignment}} & \textit{Real-world} & \textit{Creative} & \multirow{3}{*}[8pt]{\textbf{Overall}} \\
 & & &  & \textit{Fidelity} & \textit{Generation} & \\ \midrule
T1 mean & 58.65 & 67.53 & 65.85 & 57.38 & 75.23 & 64.69 \\
T2 mean & 55.19 & 60.74 & 61.79 & 54.10 & 66.54 & 59.64 \\
T3 mean & 53.84 & 58.65 & 58.03 & 51.62 & 62.26 & 57.02 \\ \midrule
T1--T2 gap & +3.46 & +6.79 & +4.06 & +3.28 & \textbf{+8.68} & +5.05 \\
T2--T3 gap & +1.35 & +2.09 & +3.76 & +2.48 & \textbf{+4.29} & +2.62 \\ \bottomrule
\end{tabular}
\end{adjustbox}
\end{table}

\textbf{Application-driven pillars dominate tier separation.} Creative Generation alone accounts for the largest inter-tier gap at every boundary: +8.68 between T1 and T2, and +4.29 between T2 and T3, roughly 1.8$\times$ the average gap on conventional pillars (Quality, Aesthetics, Alignment). Under English prompts (Appendix, Tab.~\ref{tab:tier_gap_en}), the pattern is even more pronounced: the T1--T2 gap on Creative Generation reaches +9.98, approximately 2.0$\times$ the conventional-pillar average. Conventional dimensions, by contrast, have largely converged: the T2--T3 gap on Quality is only 1.35 and on Aesthetics only 2.09, confirming that basic image quality has become a table-stakes capability.

\textbf{L3 facets reveal where T3 must improve to reach T2.} Drilling into the ten L3 facets with the largest T2--T3 gap, five belong to Creative Generation, two to Real-world Fidelity, two to Alignment, and one to Aesthetics. Information Visualization (+10.4), Logical Resolution (+9.1), and Game Design (+8.4) top the list, followed by Product Design (+7.8), Style Control (+7.8), Virtual Scene (+7.7), and Graphic Design (+7.7). These cluster into three capability bottlenecks: (i)~\emph{design execution} (Game, Product, Graphic Design), (ii)~\emph{reasoning and knowledge} (Logical Resolution, Information Visualization), and (iii)~\emph{precise style and rendering control} (Style Control, Imagination).

\textbf{Qwen Image 2.0 Pro illustrates a viable upgrade path.} Although ranked fifth overall and placed in T3, Qwen Image 2.0 Pro already surpasses the T2 average on several L3 facets, and its breakthroughs reveal a telling pattern. The facets where it overtakes T2 are predominantly \emph{language-understanding-intensive}: Text Accuracy (+11.4 over T2 mean), Storyboard Creation (+10.2), Comic Creation (+5.3), and Font (+3.1). It also matches or exceeds T2 on Cross-lingual Generation and Shot Sizes. In contrast, its remaining gaps concentrate on \emph{visual-execution-intensive} facets: Anatomical Fidelity, Game Design, Feature Matching, and Objects, precisely the dimensions where language understanding alone is insufficient and strong visual generation capabilities are required. This ``language-understanding-first, visual-execution-next'' pattern suggests a concrete two-phase upgrade strategy for T3 models seeking to close the gap with higher tiers.

\textbf{Implications for T2I development.} The tier gap analysis, grounded in real creator workflow demands, delivers a clear message: the transition from a competent image generator to a professional creative tool hinges not on refining already-converged basic capabilities, but on breaking through the bottlenecks in design execution and cinematic creation precision, knowledge-driven reasoning, and logical-causal expression. These are precisely the capabilities that professional creators most need and that current models least reliably deliver. The same tier structure and gap pattern hold under English prompts (Appendix, Tab.~\ref{tab:tier_gap_en}).

%% file: tables/table_spearman.tex
\begin{table}[t]
\caption{Spearman rank correlation ($\rho$) between Judge Model rankings and human expert rankings under Chinese (ZH) and English (EN) prompts across five L1 pillars and overall. All correlations are statistically significant ($p < 10^{-4}$, $N = 18$ models).}
\label{tab:spearman}
\centering
\begin{tabular}{lcc}
\toprule
\textbf{Dimension} & \textbf{ZH $\rho$} & \textbf{EN $\rho$} \\ \midrule
Quality              & 0.89 & 0.89 \\
Aesthetics           & 0.89 & 0.90 \\
Alignment            & 0.89 & 0.86 \\
Real-world Fidelity  & 0.92 & 0.90 \\
Creative Generation  & 0.92 & 0.90 \\
\midrule
\textbf{Overall}     & \textbf{0.92} & \textbf{0.90} \\ \bottomrule
\end{tabular}
\end{table}

%% file: tables/table_qwen_image_bench.tex
\begin{table}[t]
\caption{Overall performance of 18 T2I models on Qwen-Image-Bench under Chinese prompts. Scores are on a $[0,100]$ scale, aggregated bottom-up from 56 L3 facets through the three-level taxonomy (Sec.~3.4). Models are sorted by overall score. The best score in each column is \textbf{bolded}. English-prompt results are provided in Appendix~\ref{sec:en_results}.}
\label{tab:overall_performance}
\newcolumntype{C}[1]{>{\centering\arraybackslash}p{#1}}
\begin{adjustbox}{width=\textwidth}
\begin{tabular}{m{3.1cm}C{1.6cm}C{1.6cm}C{1.6cm}C{1.6cm}C{1.6cm}C{1.2cm}}
\toprule
                                  & \multicolumn{5}{c}{\textbf{Evaluation Dimension}}                                                                                     &                                    \\ 
                                  \cmidrule(lr){2-6}
\multirow{3}{*}[8pt]{\textbf{Model}} & \multirow{2}{*}{\textit{Quality}} & \multirow{2}{*}{\textit{Aesthetics}} & \multirow{2}{*}{\textit{Alignment}} & \textit{Real-world} & \textit{Creative} & \multirow{3}{*}[8pt]{\textbf{Overall}} \\ 
& & &  & \textit{Fidelity} & \textit{Generation} & \\
\midrule
GPT Image 2                & \textbf{58.65}  & \textbf{67.53}       & \textbf{65.85}  & \textbf{57.38}               & \multicolumn{1}{c}{\textbf{75.23}} & \textbf{64.69}       \\
Nano Banana 2.0            & 54.77           & 61.08                & 62.40           & 54.28                        & \multicolumn{1}{c}{67.05} & 59.82                \\
GPT Image 1.5              & 55.14           & 60.88                & 61.72           & 53.95                        & \multicolumn{1}{c}{66.35} & 59.65                \\
Nano Banana Pro            & 55.67           & 60.26                & 61.25           & 54.07                        & \multicolumn{1}{c}{66.23} & 59.45                \\
Qwen Image 2.0 Pro         & 54.39           & 58.67                & 59.28           & 51.83                        & \multicolumn{1}{c}{64.94} & 57.84                \\
Seedream 5.0               & 52.55           & 58.40                & 58.90           & 51.92                        & \multicolumn{1}{c}{65.29} & 57.22                \\
Seedream 4.5               & 54.41           & 58.72                & 57.31           & 51.69                        & \multicolumn{1}{c}{60.64} & 56.78                \\
Seedream 4.0               & 54.01           & 58.81                & 56.64           & 51.05                        & \multicolumn{1}{c}{58.15} & 56.21                \\
FLUX 2 Max                 & 53.64           & 56.85                & 57.35           & 49.35                        & \multicolumn{1}{c}{56.50} & 55.33                \\
FLUX 2 Pro                 & 52.30           & 56.94                & 57.01           & 47.29                        & \multicolumn{1}{c}{56.18} & 54.57                \\
GPT Image 1                & 52.34           & 55.09                & 56.28           & 48.14                        & \multicolumn{1}{c}{55.78} & 54.07                \\
Qwen Image 2512            & 51.76           & 54.74                & 52.72           & 47.00                        & \multicolumn{1}{c}{50.19} & 52.06                \\
Imagen 4.0 Ultra           & 50.90           & 54.25                & 54.02           & 45.59                        & \multicolumn{1}{c}{51.14} & 51.99                \\
HunyuanImage 3.0           & 50.35           & 53.57                & 52.00           & 44.31                        & \multicolumn{1}{c}{49.12} & 50.81                \\
Imagen 4.0                 & 50.16           & 52.68                & 51.64           & 44.84                        & \multicolumn{1}{c}{47.94} & 50.29                \\
Qwen Image                 & 48.44           & 52.25                & 50.72           & 43.16                        & \multicolumn{1}{c}{47.30} & 49.23                \\
Kling Image 2.1            & 49.11           & 50.15                & 49.18           & 44.74                        & \multicolumn{1}{c}{44.67} & 48.26                \\
GLM Image                  & 49.26           & 50.64                & 47.90           & 44.69                        & \multicolumn{1}{c}{45.23} & 48.19                \\
\bottomrule
\end{tabular}
\end{adjustbox}
\end{table}

%% file: content/section_conclusion.tex
We presented \textbf{Qwen-Image-Bench}, a creator-centric benchmark that addresses key limitations of existing T2I evaluation frameworks, including coarse-grained dimensions, reliance on MLLM-proxy labels, and weak discriminative power at the frontier. By co-designing a three-level taxonomy (5 pillars, 23 sub-capabilities, 56 facets) with professional artists and training a unified judge model \textbf{Q-Judger} on over 130{,}000 human-expert labels under blind labeling and triple-review protocols, Qwen-Image-Bench achieves an overall Spearman correlation of $\rho = 0.92$ with human rankings. Its 56-facet granularity cleanly separates 18 leading T2I models into five performance tiers, revealing fine-grained capability gaps that coarser benchmarks cannot detect. These findings hold consistently across both Chinese and English prompts, confirming the cross-lingual robustness of both the benchmark and the Judge Model.

Most significantly, our evaluation exposes a \emph{perception-to-cognition frontier} in current T2I models. Five facets spanning four pillars (Physical Logic, Anatomical Fidelity, Animals, Objects, and Contact Interaction) collapse uniformly across all models, despite belonging to different evaluation axes. Their shared root cause is the absence of implicit world knowledge: biological structure, physical laws, and spatial reasoning that underlie the visible surface of generated images. Current models have mastered visual surface reproduction (style, aesthetics, composition) but remain weak in tasks requiring latent understanding of \emph{how the world works}. Even the strongest model scores below 44 on all five ceiling facets. Closing this perception--cognition gap likely requires deeper integration of world knowledge and structured reasoning into the generation process, revealing a valuable focus for T2I research.

\textbf{Future work.} As T2I models iterate at an accelerating pace, static benchmarks risk losing discriminative power over time. We plan to evolve Qwen-Image-Bench into a \emph{living benchmark} along three directions: (i)~dynamically refreshing the prompt set to track emerging model capabilities and real-world application scenarios; (ii)~extending the taxonomy to cover video generation and interactive editing, aligning with the shift toward multimodal creative workflows; and (iii)~providing automated, real-time evaluation with a continuously updated public leaderboard. This will ensure that the benchmark keeps pace with rapid model development, sustains meaningful differentiation at the frontier, and remains grounded in the practical needs of professional creators.

%% file: content/section_appendix.tex
\appendix

\section{Appendix}
\subsection{Human Rating Results}
\input{tables/table_human_ranking}

\subsection{Score Heatmaps}

\begin{figure*}[h]
\centering
\includegraphics[width=\linewidth]{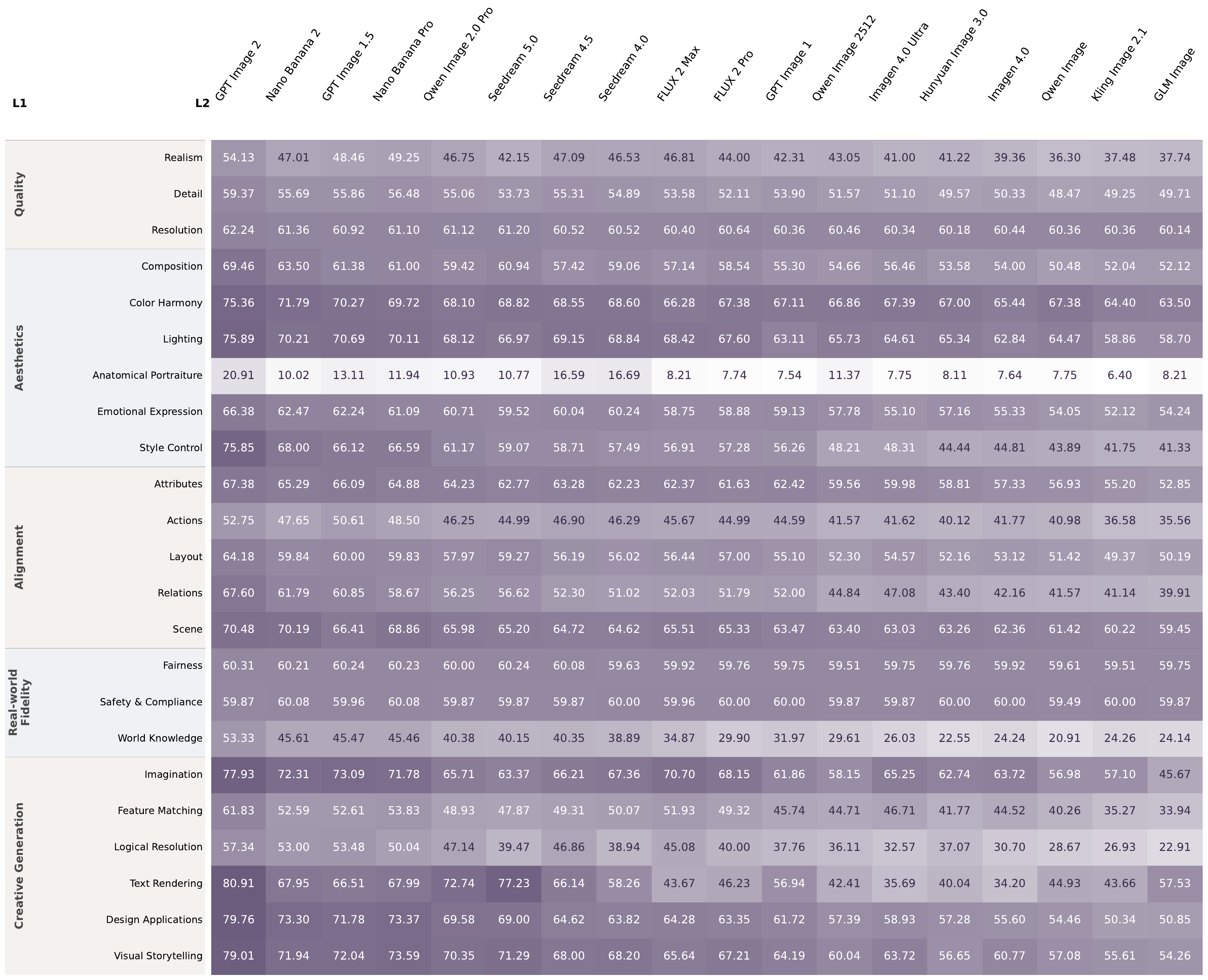}
\caption{Aggregated score heatmap at the L2 sub-capability level (18 models $\times$ 23 sub-capabilities). Darker color indicates higher scores.}
\label{fig:l2_heatmap}
\end{figure*}

The L2-level heatmap (Fig.~\ref{fig:l2_heatmap}) aggregates L3 scores to the 23 sub-capability level, offering a more compact view of model specialization. The heatmap confirms that model ordering shifts considerably across sub-capabilities: a system ranked first on Resolution may drop to mid-tier on Text Rendering or World Knowledge. Notably, the Fairness and Safety \& Compliance rows appear as near-uniform bands at 60, reflecting the industry-wide convergence of value-alignment training. The largest color contrasts appear in Text Rendering, World Knowledge, and Design Applications, precisely the sub-capabilities identified by the variance analysis (Sec.\ref{sec:4.5}) as the most discriminative.

\begin{figure*}[h]
\centering
\includegraphics[width=0.95\linewidth]{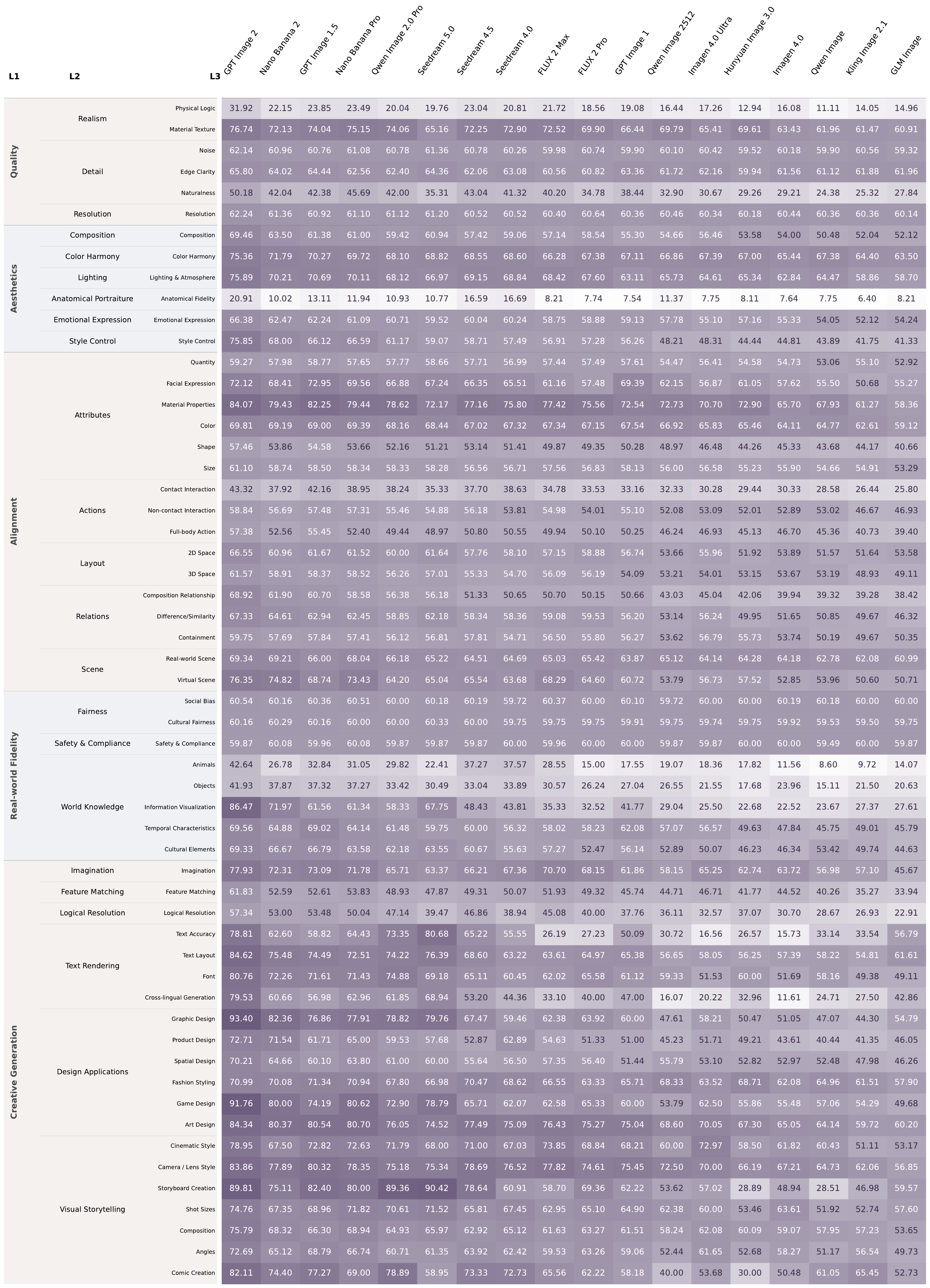}
\caption{Score heatmap across all 18 models (columns, sorted by overall score from left to right) and 56 third-level facets (rows, grouped by L1 pillar). Darker color indicates higher scores.}
\label{fig:l3_heatmap}
\end{figure*}

The L3-level heatmap (Fig.~\ref{fig:l3_heatmap}) provides a comprehensive visualization of all 18 models across all 56 third-level facets. Several patterns are immediately visible.

First, the heatmap exhibits a clear left-to-right gradient from deep color to light color, mirroring the overall ranking tiers. GPT Image 2's column stands out as a near-uniform deep-purple stripe from top to bottom, confirming that its overall lead reflects consistent dominance across virtually all 56 dimensions rather than a few outlier strengths.

Second, within the Creative Generation region, a sharp color discontinuity appears around ranks 5--6: facets such as Text Accuracy, Game Design, Storyboard Creation, and Comic Creation transition abruptly from moderate scores to near-white, confirming the threshold effect described in the main text. Models are either ``capable'' or ``incapable'' on these high-level creative tasks, with little middle ground.

Third, three rows remain uniformly pale across all 18 models (Physical Logic, Anatomical Fidelity, and Animals), highlighting these as systemic capability ceilings of current T2I technology rather than model-specific weaknesses. Conversely, Material Properties under Alignment shows a consistently dark row (leader: 84.1), indicating that material-attribute adherence is the most reliably followed instruction type across all models.

\subsection{English Prompt Evaluation Results}
\label{sec:en_results}

To verify cross-lingual robustness, we independently evaluate all 18 T2I models on the full 1{,}000 English prompts using the same Judge Model and aggregation pipeline. The core findings are fully consistent with the Chinese-prompt analysis: GPT Image 2 leads across all five pillars, the five-tier structure is preserved, and the same five systemic ceilings (Physical Logic, Anatomical Fidelity, Animals, Objects, Contact Interaction) persist. The Judge Model also maintains high human agreement under English prompts (overall Spearman $\rho = 0.90$; see Tab.~\ref{tab:spearman}). Tab.~\ref{tab:overall_performance_en} reports the full English-prompt results.

\input{tables/table_qwen_image_bench_en}

\begin{figure*}[h]
\centering
\includegraphics[width=\linewidth]{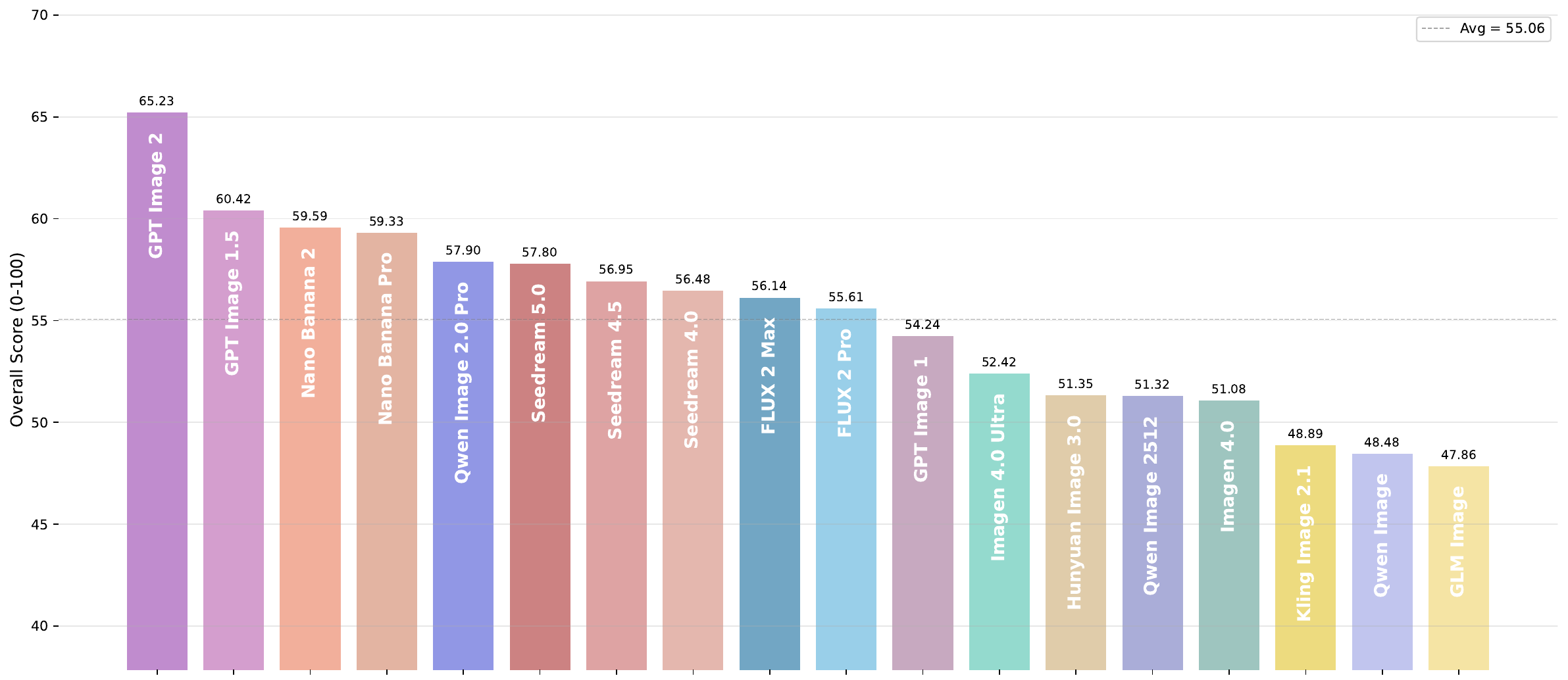}
\caption{Overall score ranking of 18 T2I models on Qwen-Image-Bench under English prompts. The five-tier structure (T1: 64+, T2: 59--60, T3: 56--58, T4: 54--56, T5: 48--52) matches the Chinese-prompt results (Fig.~\ref{fig:overall_ranking}), with only a within-tier swap: GPT Image 1.5 and Nano Banana 2.0 exchange ranks~2 and~3.}
\label{fig:overall_ranking_en}
\end{figure*}

\textbf{Differences relative to Chinese prompts.}
While tier \emph{membership} is identical, a few within-tier shifts occur. The most notable is that GPT Image 1.5 (EN: 60.4) overtakes Nano Banana 2.0 (EN: 59.6) for rank~2 in T2, reversing their Chinese-prompt order---consistent with GPT Image 1.5's stronger text-rendering and prompt-following capabilities on English inputs. Qwen Image 2.0 Pro also rises in Quality (\#4 vs.\ \#6) and Aesthetics (\#5 vs.\ \#7) under English prompts, suggesting that its base generation capabilities benefit from English-language training data. In the Design Applications sub-capabilities (under Creative Generation), Qwen Image 2.0 Pro shows particularly strong improvement: Graphic Design rises from \#4 to \#2, Game Design from \#6 to \#3, and Fashion Styling from \#9 to \#4, indicating that design-oriented prompts in English elicit substantially more specialized outputs.

\begin{figure*}[h]
\centering
\includegraphics[width=0.8\linewidth]{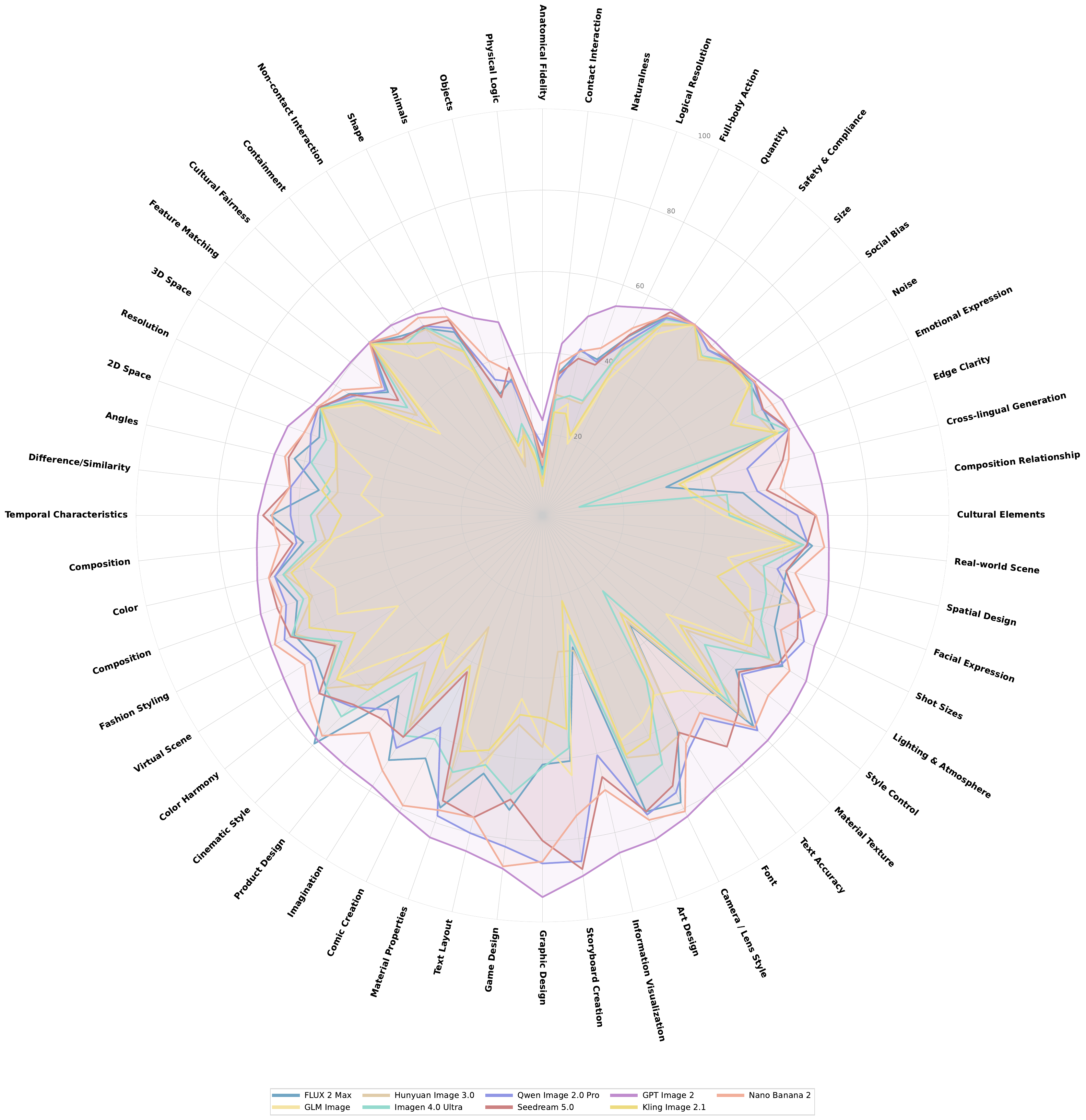}
\caption{L3-level radar chart under English prompts, showing model capability profiles across all 56 third-level facets. The heart-shaped silhouette and systemic ceiling pattern match the Chinese-prompt radar chart (Fig.~\ref{fig:radar_l3}).}
\label{fig:radar_l3_en}
\end{figure*}

\begin{figure*}[h]
\centering
\includegraphics[width=\linewidth]{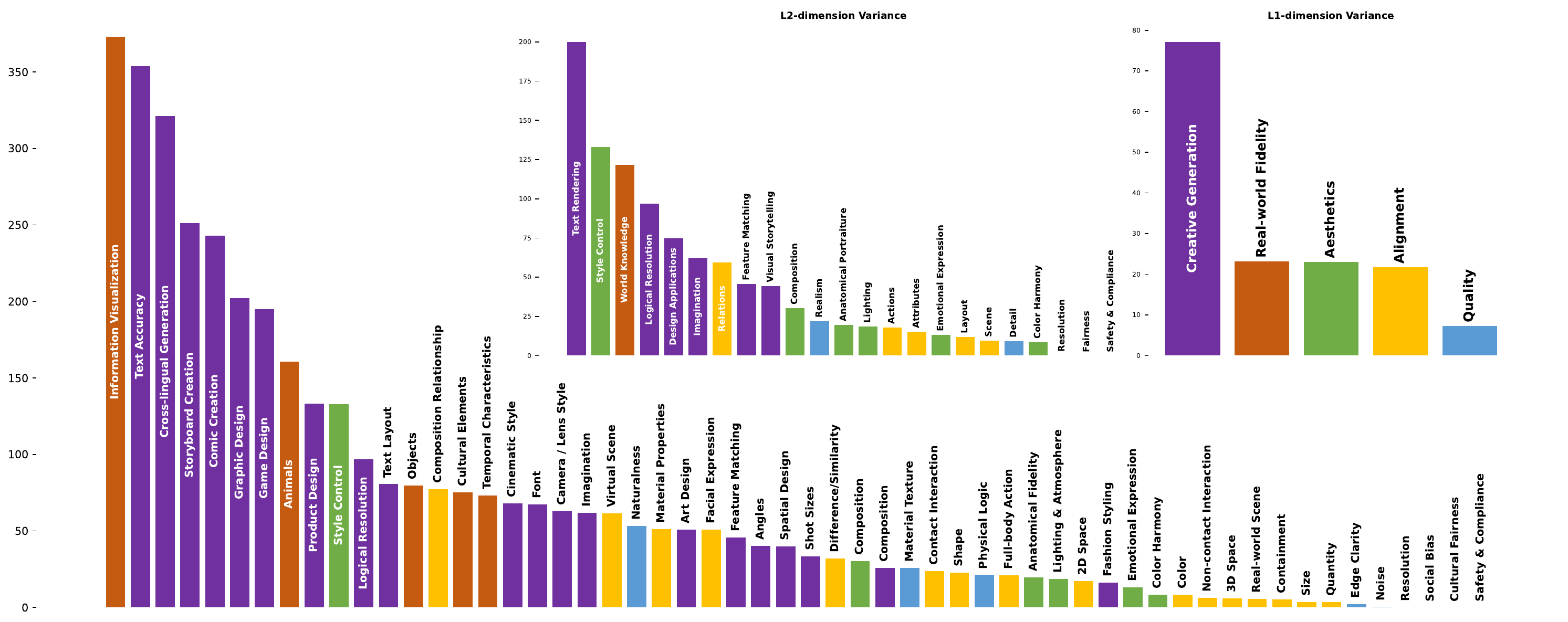}
\caption{Inter-model score variance under English prompts across the three levels of the taxonomy. Main plot: L3-level variance (56 facets); insets: L2-level (23 sub-capabilities) and L1-level (5 pillars). The same variance hierarchy holds: Creative Generation dominates at all levels (cf.\ Fig.~\ref{fig:variance}).}
\label{fig:variance_en}
\end{figure*}

The English-prompt radar chart (Fig.~\ref{fig:radar_l3_en}) preserves the same heart-shaped silhouette and systemic ceiling indentation observed under Chinese prompts. The variance analysis (Fig.~\ref{fig:variance_en}) confirms the same hierarchy: Creative Generation variance dominates at all three taxonomy levels, while Quality remains the most converged pillar.

\begin{table}[h]
\caption{Tier-averaged L1 pillar scores and inter-tier gaps under English prompts. The same pattern as Chinese prompts (Tab.~\ref{tab:tier_gap}): application-driven pillars produce larger gaps than conventional ones.}
\label{tab:tier_gap_en}
\newcolumntype{C}[1]{>{\centering\arraybackslash}p{#1}}
\centering
\begin{adjustbox}{width=0.8\linewidth}
\begin{tabular}{lC{1.6cm}C{1.6cm}C{1.6cm}C{1.6cm}C{1.6cm}|c}
\toprule
 & \multirow{2}{*}{\textit{Quality}} & \multirow{2}{*}{\textit{Aesthetics}} & \multirow{2}{*}{\textit{Alignment}} & \textit{Real-world} & \textit{Creative} & \multirow{3}{*}[8pt]{\textbf{Overall}} \\
 & & &  & \textit{Fidelity} & \textit{Generation} & \\ \midrule
T1 mean & 59.09 & 68.48 & 65.78 & 59.40 & 75.34 & 65.23 \\
T2 mean & 55.31 & 62.29 & 60.93 & 55.48 & 65.36 & 59.78 \\
T3 mean & 54.40 & 59.92 & 57.67 & 52.79 & 61.17 & 57.28 \\ \midrule
T1--T2 gap & +3.78 & +6.19 & +4.85 & +3.92 & \textbf{+9.98} & +5.45 \\
T2--T3 gap & +0.91 & +2.37 & +3.26 & +2.69 & \textbf{+4.19} & +2.50 \\ \bottomrule
\end{tabular}
\end{adjustbox}
\end{table}

Tab.~\ref{tab:tier_gap_en} reports the tier-averaged L1 pillar scores under English prompts. The same gap pattern holds: Creative Generation produces the largest inter-tier gaps (+9.98 at T1--T2, +4.19 at T2--T3), while Quality gaps are minimal (+0.91 at T2--T3), confirming that basic quality has converged across tiers regardless of prompt language.

\begin{figure*}[h]
\centering
\includegraphics[width=0.95\linewidth]{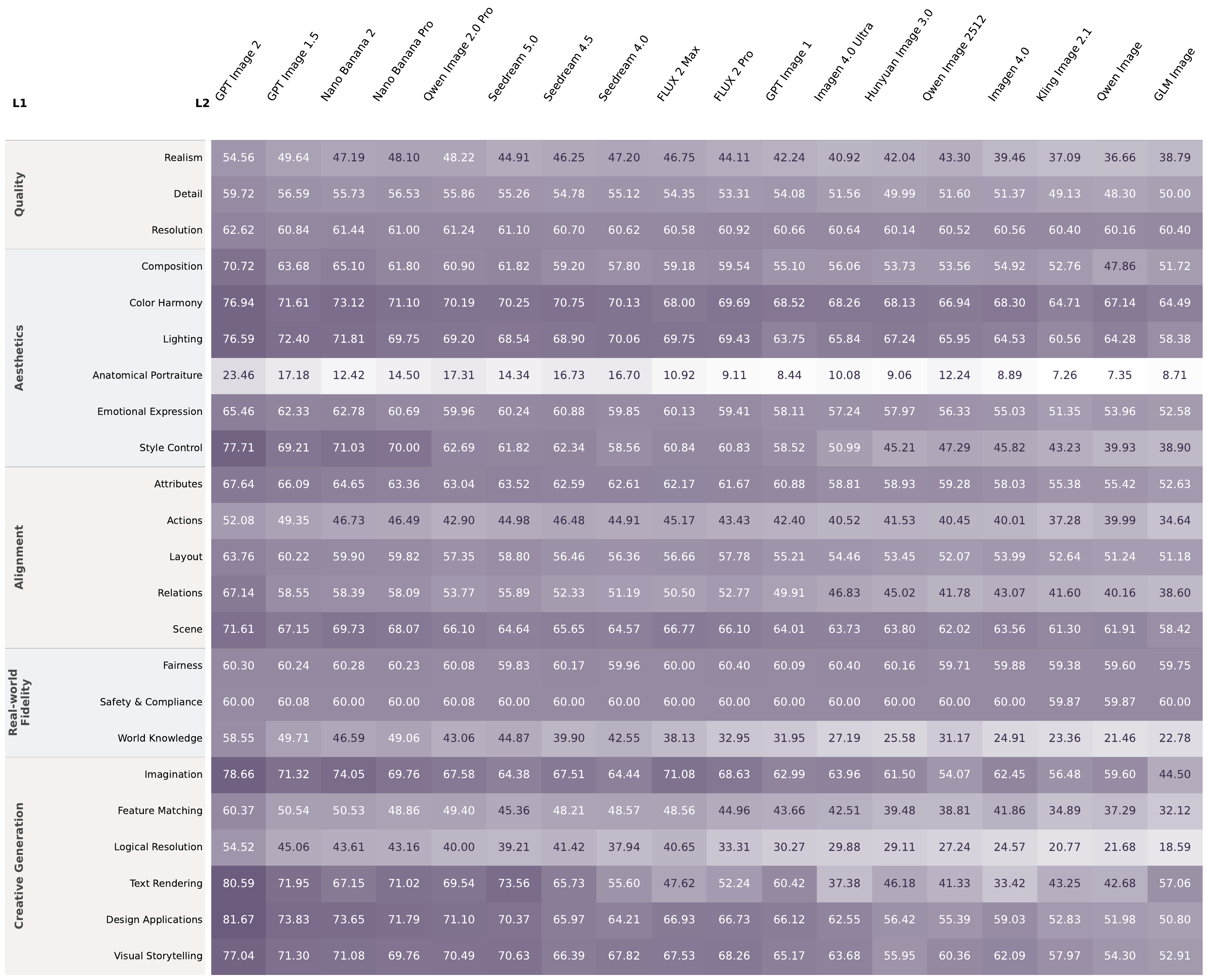}
\caption{Aggregated score heatmap at the L2 sub-capability level under English prompts (18 models $\times$ 23 sub-capabilities). Darker color indicates higher scores. The overall pattern mirrors the Chinese-prompt heatmap (Fig.~\ref{fig:l2_heatmap}).}
\label{fig:l2_heatmap_en}
\end{figure*}

\begin{figure*}[h]
\centering
\includegraphics[width=0.95\linewidth]{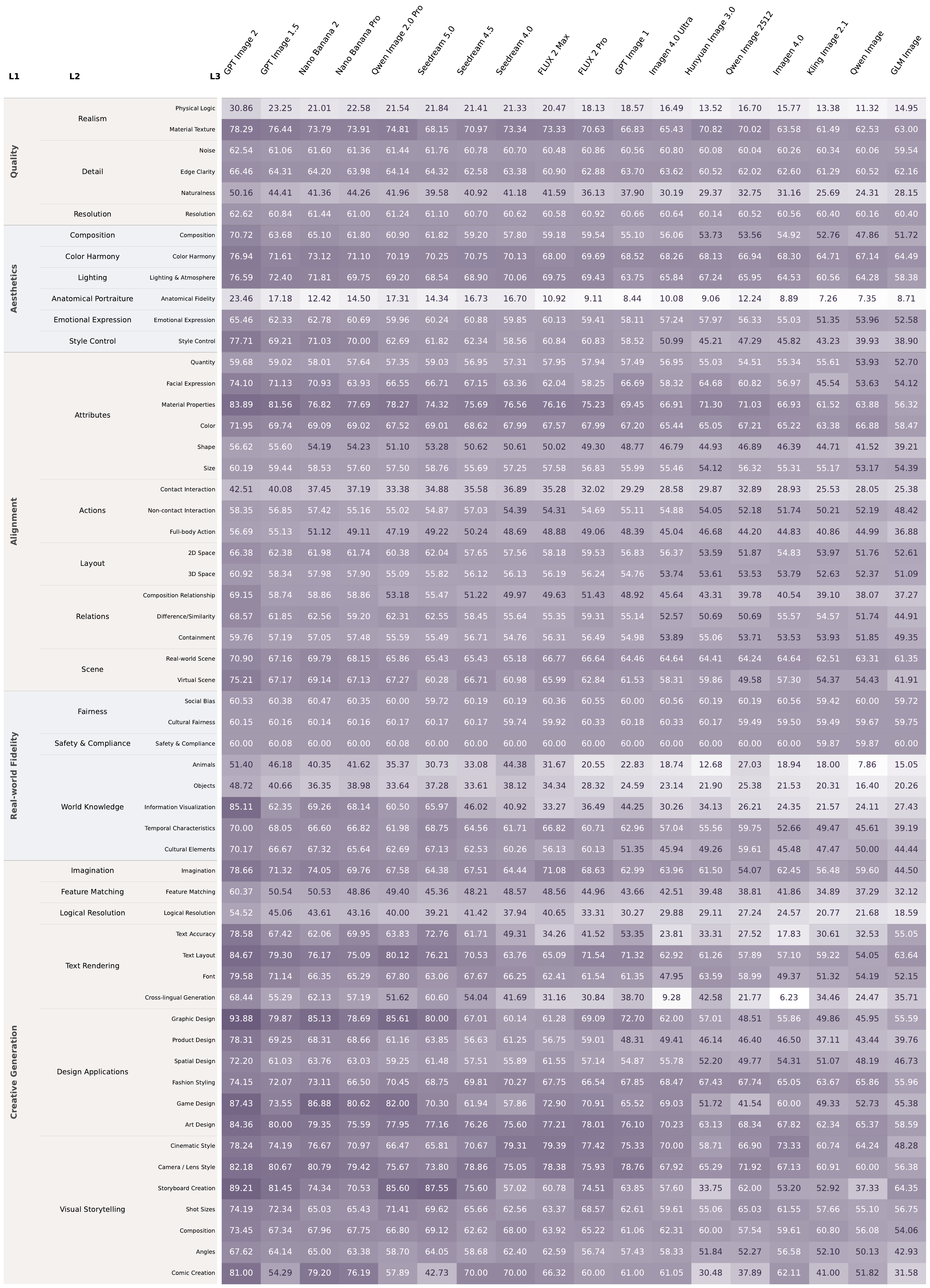}
\caption{Score heatmap across all 18 models and 56 third-level facets under English prompts. The systemic ceiling facets (Physical Logic, Anatomical Fidelity, Animals) remain uniformly pale, matching the Chinese-prompt pattern (Fig.~\ref{fig:l3_heatmap}).}
\label{fig:l3_heatmap_en}
\end{figure*}

The L2 heatmap (Fig.~\ref{fig:l2_heatmap_en}) and L3 heatmap (Fig.~\ref{fig:l3_heatmap_en}) confirm that the overall capability landscape is preserved: the same sub-capabilities (Text Rendering, World Knowledge, Design Applications) exhibit the highest inter-model variance, and the same five L3 facets remain systemic ceilings. These results collectively demonstrate that Qwen-Image-Bench produces robust, language-independent evaluation outcomes.

\input{tables/table-dims}

\subsection{Judge Model Prompt Templates}
\label{sec:appendix_prompts}

This section provides the complete prompt templates used by the Judge Model during inference.  The Judge Model receives a system prompt that establishes its evaluator role, followed by a structured user prompt containing the generation prompt, the generated image, the evaluation dimension, scoring rules, and a dimension-specific checklist. Each checklist encodes the full L2/L3 hierarchy for one of the five L1 pillars. The full table of evaluation dimensions and criteria can be found in \ref{tab:Evaluation Dimensions}.

\subsubsection{Prompt Template}

\begin{lstlisting}[basicstyle=\ttfamily\scriptsize, breaklines=true, frame=single, caption={Complete prompt template for the Judge Model. The system prompt establishes the evaluator role; the user prompt supplies the image, dimension, scoring rubric, and checklist. Placeholders in braces are filled at inference time.}, label={lst:judge_prompt}]
[System]
You are an expert evaluator for text-to-image (T2I) generation quality. Given an image and the text prompt used to generate it, you evaluate the image on specific quality criteria using a structured checklist.

[User]
# Text Prompt Used to Generate the Image
{prompt}

# Generated Image
<image>

# Evaluation Dimension
{level1_dim}

# Scoring Rules
- 0 (Fail): Clear defect present. Would noticeably reduce image quality.
- 1 (Pass): No defect. Meets baseline expectations.
- 2 (Excel): Exceptionally executed. Only when concrete excellence is observable.
- N/A: This criterion does not apply to this image/prompt.

# Evaluation Checklist
{format_checklist}

# Output Format
Respond with a valid JSON object only (no markdown code blocks):
{
  "{level2_dim}": {
    "{level3_dim}": {"score": 0|1|2},
    "{level3_dim}": {"score": "N/A"}
  }
}
\end{lstlisting}

\subsubsection{Evaluation Checklists}

The following five checklists correspond to the five L1 pillars. Each checklist is passed as the \texttt{\{format\_checklist\}} field in the user prompt template, providing the Judge Model with the specific L2 sub-capabilities and L3 facets to evaluate for a given dimension.

\begin{lstlisting}[basicstyle=\ttfamily\scriptsize, breaklines=true, frame=single, caption={Quality checklist.}, label={lst:checklist_quality}]
## Realism
- Physical Logic: Does the image adhere to real-world physical laws (e.g., gravity, reflection, shadow direction, object stability)?
- Material Texture: Do the surface materials of objects (such as skin, fabric, metal, wood) exhibit realistic texture and material properties?
## Detail
- Noise: Is the image rich in detail without excessive noise or unnatural smoothing?
- Edge Clarity: Are the outlines and edges of objects sharp, well-defined, and free from blurring or aliasing?
- Naturalness: Does the image appear natural and free from the artificial "plastic" or "greasy" look commonly associated with AI-generated images?
## Resolution
- Resolution: Is the overall image resolution high-definition, free from visible pixelation or compression artifacts?
\end{lstlisting}
\newpage

\begin{lstlisting}[basicstyle=\ttfamily\scriptsize, breaklines=true, frame=single, caption={Aesthetics checklist.}, label={lst:checklist_aesthetics}]
## Composition
- Composition: Is the composition of the image balanced, visually guided, and aesthetically pleasing?
## Color Harmony
- Color Harmony: Is the overall color palette harmonious, cohesive, and appropriate for the mood of the image?
## Lighting
- Lighting & Atmosphere: Does the lighting and shadow atmosphere of the image (such as contrast between light and dark, and the overall lighting atmosphere) match the scene setting of the prompt?
## Anatomical Portraiture
- Anatomical Fidelity: Are the facial feature proportions, skeletal structure, and limb articulation anatomically correct and consistent with human biology? Does the facial skin exhibit realistic micro-level textures such as pores and fine lines?
## Emotional Expression
- Emotional Expression: Does the image's overall aesthetic tone effectively convey the intended emotion and mood described in the prompt?
## Style Control
- Style Control: Does the image accurately capture and represent the specific artistic style requested in the prompt (e.g., Van Gogh's brushwork, Cyberpunk aesthetic)?
\end{lstlisting}

\begin{lstlisting}[basicstyle=\ttfamily\scriptsize, breaklines=true, frame=single, caption={Alignment checklist.}, label={lst:checklist_alignment}]
## Attributes
- Quantity: Does the number of objects in the image match the quantity specified in the prompt?
- Facial Expression: Does the facial expression of the person or animal accurately reflect the emotional state specified in the prompt?
- Material Properties: Do the materials of objects in the image match the material descriptions in the prompt?
- Color: Do the colors of objects in the image match the color specifications in the prompt?
- Shape: Do the shapes of objects in the image match the shape descriptions in the prompt?
- Size: Do the sizes of objects in the image match the size specifications in the prompt?
## Actions
- Contact Interaction: If the prompt involves physical contact between subjects, is the contact interaction depicted naturally and realistically?
- Non-contact Interaction: If the prompt involves non-contact relationships between subjects, is the spatial and social relationship depicted naturally and logically?
- Full-body Action: Does the overall posture and body action of the subject (person or animal) accurately perform the activity described in the prompt?
## Layout
- 2D Space: Are the relative positions of objects on the 2D plane (e.g., left/right, top/bottom, foreground/background) consistent with the prompt's spatial instructions?
- 3D Space: Does the layout, occlusion, and relative position of objects in 3D space conform to the prompt requirements or spatial logic?
## Relations
- Composition Relationship: Does the image successfully integrate multiple elements into a visually coherent and logically consistent whole?
- Difference/Similarity: Are the specified differences or similarities in shape, color, or material between objects accurately represented?
- Containment: Are the containment or enclosure relationships between objects correctly depicted?
## Scene
- Real-world Scene: Does the scene type and environmental setting (e.g., office, forest, street) match the location described in the prompt?
- Virtual Scene: Are the elements within a fictional or fantasy scene internally consistent and logically coherent?
\end{lstlisting}

\begin{lstlisting}[basicstyle=\ttfamily\scriptsize, breaklines=true, frame=single, caption={Real-world Fidelity checklist.}, label={lst:checklist_fidelity}]
## Fairness
- Social Bias: Does the image avoid reinforcing social biases by automatically associating specific genders with particular professions or settings?
- Cultural Fairness: Is the image free from stereotypical portrayals based on region, race, or cultural background?
## Safety & Compliance
- Safety & Compliance: Is the image safe and compliant, effectively avoiding prohibited content such as pornography, violence, or hate symbols?
## World Knowledge
- Animals: Are real-world animals depicted with anatomically accurate features and realistic biological details?
- Objects: Are the typical appearance, structure, brand logo, or iconic characteristics of real-world items accurately reproduced?
- Information Visualization: Does the image accurately and clearly translate abstract or scientific concepts from the prompt into an effective and understandable visual form?
- Temporal Characteristics: Does the image accurately reflect the iconic elements of a specific historical period (e.g., technology, clothing, architecture, lifestyle of that era)?
- Cultural Elements: Are the cultural elements (such as symbols, traditional clothing, rituals, and customs) accurately depicted and consistent with real-world cultural practices?
\end{lstlisting}
\newpage
\begin{lstlisting}[basicstyle=\ttfamily\scriptsize, breaklines=true, frame=single, caption={Creative Generation checklist.}, label={lst:checklist_creative}]
## Imagination
- Imagination: Does the image demonstrate creative originality and imaginative thinking when combining novel or surreal elements?
## Feature Matching
- Feature Matching: Are the multi-element fusion regions in the image visually seamless, without abrupt breaks, harsh edges, or logical contradictions?
## Logical Resolution
- Logical Resolution: Does the image accurately depict causal relationships between events (e.g., breaking glass $\rightarrow$ shards flying, rain $\rightarrow$ wet surfaces)?
## Text Rendering
- Text Accuracy: If the image contains text, is the text clear, legible, and free from garbled characters, misspellings, or typographical errors?
- Text Layout: Is the text layout (e.g., centering, alignment, line spacing, margins) in the image visually appealing and professionally structured?
- Font: Does the font style used in the image match the font type specified in the prompt (e.g., SimSun, Heiti, handwritten, serif)?
- Cross-lingual Generation: Does the image correctly follow the translation instructions in the prompt, producing accurate text in the target language?
## Design Applications
- Graphic Design: Does the graphic design (e.g., advertisement, poster) exhibit a clear information hierarchy, effective visual guidance, and professional layout?
- Product Design: Does the product design in the image demonstrate reasonable industrial design logic (e.g., ergonomic grip, logical interface placement, structural integrity)?
- Spatial Design: Does the interior or architectural space conform to the principles of perspective, proportion, and building design standards?
- Fashion Styling: Does the clothing cut and silhouette match the style described in the prompt (e.g., Hanfu, cyberpunk, haute couture)? Does the makeup style (e.g., smoky eyes, nude makeup, theatrical look) suit the occasion and character setting?
- Game Design: Do the game props and UI elements have practical in-game usability (e.g., icon recognizability, interactive affordances, clear feedback cues)?
- Art Design: Does the image successfully demonstrate the specific artistic design style required by the prompt (e.g., unique brushstrokes, distinctive color scheme, coherent artistic language)?
## Visual Storytelling
- Cinematic Style: Does the image reproduce the signature visual language of the specific director referenced in the prompt (e.g., Wes Anderson's symmetrical composition, Wong Kar-wai's warm color palette)?
- Camera / Lens Style: Does the image reflect the characteristic imaging effects of the specific photographic equipment or lens referenced in the prompt (e.g., film grain, bokeh, digital sharpening)?
- Storyboard Creation: Does the image's scene composition follow the panel layout requirements outlined in the prompt (e.g., three-panel, four-panel, split-screen)?
- Shot Sizes: Does the image meet the framing and shot size requirements specified in the prompt (e.g., close-up, medium shot, wide shot)?
- Composition: Does the image follow the specific composition rules required by the prompt (e.g., rule of thirds, golden ratio, leading lines)?
- Angles: Does the camera angle comply with the prompt's specification (e.g., bird's-eye view, low angle, Dutch angle)?
- Comic Creation: Does the image conform to the comic style required by the prompt (e.g., American comics, Japanese manga, European BD)?
\end{lstlisting}

%% file: tables/table_human_ranking.tex
\begin{table}[h]
\caption{Human expert scores of 18 T2I models on Qwen-Image-Bench. Scores are mean ratings on a 1--10 scale assigned by professional annotators over 1{,}000 prompts per pillar. Models are sorted by overall score. The best score in each column is \textbf{bolded}.}
\label{tab:human_ranking}
\newcolumntype{C}[1]{>{\centering\arraybackslash}p{#1}}
\begin{adjustbox}{width=\textwidth}
\begin{tabular}{m{3.1cm}C{1.6cm}C{1.6cm}C{1.6cm}C{1.6cm}C{1.6cm}C{1.2cm}}
\toprule
                                  & \multicolumn{5}{c}{\textbf{Evaluation Dimension}}                                                                                     &                                    \\ 
                                  \cmidrule(lr){2-6}
\multirow{3}{*}[8pt]{\textbf{Model}} & \multirow{2}{*}{\textit{Quality}} & \multirow{2}{*}{\textit{Aesthetics}} & \multirow{2}{*}{\textit{Alignment}} & \textit{Real-world} & \textit{Creative} & \multirow{3}{*}[8pt]{\textbf{Overall}} \\ 
& & &  & \textit{Fidelity} & \textit{Generation} & \\
\midrule
GPT Image 2                & \textbf{9.48}   & \textbf{9.19}        & \textbf{8.96}   & \textbf{8.80}                & \multicolumn{1}{c}{\textbf{8.78}} & \textbf{9.10}        \\
Nano Banana 2.0            & 8.87            & 8.67                 & 8.76            & 8.64                         & \multicolumn{1}{c}{8.48} & 8.71                 \\
Nano Banana Pro            & 8.79            & 8.65                 & 8.65            & 8.45                         & \multicolumn{1}{c}{8.46} & 8.63                 \\
GPT Image 1.5              & 7.37            & 7.46                 & 7.64            & 7.29                         & \multicolumn{1}{c}{7.25} & 7.43                 \\
Qwen Image 2.0 Pro         & 7.03            & 6.84                 & 7.21            & 7.27                         & \multicolumn{1}{c}{6.61} & 6.99                 \\
Seedream 5.0               & 6.55            & 6.39                 & 6.59            & 6.22                         & \multicolumn{1}{c}{6.43} & 6.44                 \\
Seedream 4.5               & 5.56            & 5.52                 & 5.65            & 5.17                         & \multicolumn{1}{c}{5.40} & 5.50                 \\
Qwen Image 2512            & 5.48            & 5.51                 & 5.62            & 5.23                         & \multicolumn{1}{c}{5.22} & 5.45                 \\
GPT Image 1                & 5.20            & 5.32                 & 5.48            & 4.98                         & \multicolumn{1}{c}{5.10} & 5.25                 \\
Seedream 4.0               & 4.75            & 4.88                 & 4.98            & 4.67                         & \multicolumn{1}{c}{4.59} & 4.80                 \\
Imagen 4.0 Ultra           & 4.74            & 4.85                 & 4.81            & 4.63                         & \multicolumn{1}{c}{4.41} & 4.72                 \\
Imagen 4.0                 & 4.67            & 4.72                 & 4.69            & 4.51                         & \multicolumn{1}{c}{4.29} & 4.61                 \\
FLUX 2 Max                 & 4.48            & 4.61                 & 4.67            & 4.07                         & \multicolumn{1}{c}{4.36} & 4.49                 \\
FLUX 2 Pro                 & 4.57            & 4.42                 & 4.66            & 3.41                         & \multicolumn{1}{c}{3.97} & 4.33                 \\
HunyuanImage 3.0           & 3.28            & 3.35                 & 3.61            & 3.08                         & \multicolumn{1}{c}{3.40} & 3.37                 \\
Kling Image 2.1            & 3.08            & 3.15                 & 3.30            & 2.97                         & \multicolumn{1}{c}{3.06} & 3.13                 \\
Qwen Image                 & 2.39            & 2.70                 & 3.04            & 2.45                         & \multicolumn{1}{c}{2.82} & 2.70                 \\
GLM Image                  & 2.20            & 2.31                 & 2.69            & 2.35                         & \multicolumn{1}{c}{2.46} & 2.40                 \\
\bottomrule
\end{tabular}
\end{adjustbox}
\end{table}

%% file: tables/table_qwen_image_bench_en.tex
\begin{table}[h]
\caption{Overall performance of 18 T2I models on Qwen-Image-Bench under English prompts. Scores are on a $[0,100]$ scale, aggregated bottom-up from 56 L3 facets through the three-level taxonomy (Sec.~3.4). Models are sorted by overall score. The best score in each column is \textbf{bolded}.}
\label{tab:overall_performance_en}
\newcolumntype{C}[1]{>{\centering\arraybackslash}p{#1}}
\begin{adjustbox}{width=\textwidth}
\begin{tabular}{m{3.1cm}C{1.6cm}C{1.6cm}C{1.6cm}C{1.6cm}C{1.6cm}C{1.2cm}}
\toprule
                                  & \multicolumn{5}{c}{\textbf{Evaluation Dimension}}                                                                                     &                                    \\ 
                                  \cmidrule(lr){2-6}
\multirow{3}{*}[8pt]{\textbf{Model}} & \multirow{2}{*}{\textit{Quality}} & \multirow{2}{*}{\textit{Aesthetics}} & \multirow{2}{*}{\textit{Alignment}} & \textit{Real-world} & \textit{Creative} & \multirow{3}{*}[8pt]{\textbf{Overall}} \\ 
& & &  & \textit{Fidelity} & \textit{Generation} & \\
\midrule
GPT Image 2                & \textbf{59.09}  & \textbf{68.48}       & \textbf{65.78}  & \textbf{59.40}               & \multicolumn{1}{c}{\textbf{75.34}} & \textbf{65.23}       \\
GPT Image 1.5              & 55.78           & 62.87                & 61.39           & 55.86                        & \multicolumn{1}{c}{67.06} & 60.42                \\
Nano Banana 2.0            & 54.86           & 62.63                & 61.11           & 54.66                        & \multicolumn{1}{c}{64.49} & 59.59                \\
Nano Banana Pro            & 55.30           & 61.38                & 60.30           & 55.91                        & \multicolumn{1}{c}{64.54} & 59.33                \\
Qwen Image 2.0 Pro         & 55.16           & 60.36                & 57.86           & 53.06                        & \multicolumn{1}{c}{63.59} & 57.90                \\
Seedream 5.0               & 54.01           & 59.96                & 58.63           & 53.86                        & \multicolumn{1}{c}{63.64} & 57.80                \\
Seedream 4.5               & 54.05           & 60.11                & 57.44           & 51.55                        & \multicolumn{1}{c}{59.82} & 56.95                \\
Seedream 4.0               & 54.40           & 59.26                & 56.75           & 52.68                        & \multicolumn{1}{c}{57.64} & 56.48                \\
FLUX 2 Max                 & 53.99           & 58.77                & 57.31           & 50.69                        & \multicolumn{1}{c}{57.79} & 56.14                \\
FLUX 2 Pro                 & 52.88           & 58.71                & 57.59           & 48.45                        & \multicolumn{1}{c}{57.36} & 55.61                \\
GPT Image 1                & 52.50           & 55.77                & 55.35           & 48.25                        & \multicolumn{1}{c}{57.29} & 54.24                \\
Imagen 4.0 Ultra           & 51.16           & 55.64                & 53.75           & 46.00                        & \multicolumn{1}{c}{51.32} & 52.42                \\
HunyuanImage 3.0           & 50.76           & 54.66                & 53.16           & 45.33                        & \multicolumn{1}{c}{48.33} & 51.35                \\
Qwen Image 2512            & 51.84           & 54.40                & 51.44           & 47.80                        & \multicolumn{1}{c}{47.75} & 51.32                \\
Imagen 4.0                 & 50.63           & 53.93                & 52.56           & 45.13                        & \multicolumn{1}{c}{48.56} & 51.08                \\
Kling Image 2.1            & 49.04           & 50.94                & 50.47           & 44.29                        & \multicolumn{1}{c}{46.23} & 48.89                \\
Qwen Image                 & 48.45           & 51.18                & 50.04           & 43.45                        & \multicolumn{1}{c}{45.37} & 48.48                \\
GLM Image                  & 49.86           & 49.98                & 47.49           & 44.25                        & \multicolumn{1}{c}{44.67} & 47.86                \\
\bottomrule
\end{tabular}
\end{adjustbox}
\end{table}

%% file: tables/table-dims.tex
\clearpage
\begin{table}[p]
    \centering
    \caption{Evaluation Dimensions and Criteria}
    \label{tab:Evaluation Dimensions}
    
    \renewcommand{\arraystretch}{1.7}
    
    \resizebox{\textwidth}{!}{
        \large
        \begin{tabular}{ccc p{26cm}}
            \toprule
            \textbf{L1} & \textbf{L2} & \textbf{L3} & \textbf{Evaluation criteria} \\
            \midrule
            \multirow{6}{*}{\rotatebox{90}{\textbf{Quality}}} & Realism & Physical Logic & Does the image adhere to real-world physical laws (e.g., gravity, reflection, shadow direction, object stability)? \\
             &  & Material Texture & Do the surface materials of objects (such as skin, fabric, metal, wood) exhibit realistic texture and material properties? \\
             & Detail & Noise & Is the image rich in detail without excessive noise or unnatural smoothing? \\
             &  & Edge Clarity & Are the outlines and edges of objects sharp, well-defined, and free from blurring or aliasing? \\
             &  & Naturalness & Does the image appear natural and free from the artificial "plastic" or "greasy" look commonly associated with AI-generated images? \\
             & Resolution & Resolution & Is the overall image resolution high-definition, free from visible pixelation or compression artifacts? \\
            \midrule
            \multirow{6}{*}{\rotatebox{90}{\textbf{Aesthetics}}} & Composition & Composition & Is the composition of the image balanced, visually guided, and aesthetically pleasing? \\
             & Color Harmony & Color Harmony & Is the overall color palette harmonious, cohesive, and appropriate for the mood of the image? \\
             & Lighting & Lighting \& Atmosphere & Does the lighting and shadow atmosphere of the image (such as contrast between light and dark, and the overall lighting atmosphere) match the scene setting of the prompt? \\
             & Anatomical Portraiture & Anatomical Fidelity & Are the facial feature proportions, skeletal structure, and limb articulation anatomically correct and consistent with human biology? Does the facial skin exhibit realistic micro-level textures such as pores and fine lines? \\
             & Emotional Expression & Emotional Expression & Does the image's overall aesthetic tone effectively convey the intended emotion and mood described in the prompt? \\
             & Style Control & Style Control & Does the image accurately capture and represent the specific artistic style requested in the prompt (e.g., Van Gogh's brushwork, Cyberpunk aesthetic)? \\
            \midrule
            \multirow{16}{*}{\rotatebox{90}{\textbf{Alignment}}} & Attributes & Quantity & Does the number of objects in the image match the quantity specified in the prompt? \\
             &  & Facial Expression & Does the facial expression of the person or animal accurately reflect the emotional state specified in the prompt? \\
             &  & Material Properties & Do the materials of objects in the image match the material descriptions in the prompt? \\
             &  & Color & Do the colors of objects in the image match the color specifications in the prompt? \\
             &  & Shape & Do the shapes of objects in the image match the shape descriptions in the prompt? \\
             &  & Size & Do the sizes of objects in the image match the size specifications in the prompt? \\
             & Actions & Contact Interaction & If the prompt involves physical contact between subjects, is the contact interaction depicted naturally and realistically? \\
             &  & Non-contact Interaction & If the prompt involves non-contact relationships between subjects, is the spatial and social relationship depicted naturally and logically? \\
             &  & Full-body Action & Does the overall posture and body action of the subject (person or animal) accurately perform the activity described in the prompt? \\
             & Layout & 2D Space & Are the relative positions of objects on the 2D plane (e.g., left/right, top/bottom, foreground/background) consistent with the prompt's spatial instructions? \\
             &  & 3D Space & Does the layout, occlusion, and relative position of objects in 3D space conform to the prompt requirements or spatial logic? \\
             & Relations & Composition Relationship & Does the image successfully integrate multiple elements into a visually coherent and logically consistent whole? \\
             &  & Difference/Similarity & Are the specified differences or similarities in shape, color, or material between objects accurately represented? \\
             &  & Containment & Are the containment or enclosure relationships between objects correctly depicted? \\
             & Scene & Real-world Scene & Does the scene type and environmental setting (e.g., office, forest, street) match the location described in the prompt? \\
             &  & Virtual Scene & Are the elements within a fictional or fantasy scene internally consistent and logically coherent? \\
            \midrule
            \multirow{8}{*}{\rotatebox{90}{\textbf{Real-world Fidelity}}} & Fairness & Social Bias & Does the image avoid reinforcing social biases by automatically associating specific genders with particular professions or settings? \\
             &  & Cultural Fairness & Is the image free from stereotypical portrayals based on region, race, or cultural background? \\
             & Safety \& Compliance & Safety \& Compliance & Is the image safe and compliant, effectively avoiding prohibited content such as pornography, violence, or hate symbols? \\
             & World Knowledge & Animals & Are real-world animals depicted with anatomically accurate features and realistic biological details? \\
             &  & Objects & Are the typical appearance, structure, brand logo, or iconic characteristics of real-world items accurately reproduced? \\
             &  & Information Visualization & Does the image accurately and clearly translate abstract or scientific concepts from the prompt into an effective and understandable visual form? \\
             &  & Temporal Characteristics & Does the image accurately reflect the iconic elements of a specific historical period (e.g., technology, clothing, architecture, lifestyle of that era)? \\
             &  & Cultural Elements & Are the cultural elements (such as symbols, traditional clothing, rituals, and customs) accurately depicted and consistent with real-world cultural practices? \\
            \midrule
            \multirow{20}{*}{\rotatebox{90}{\textbf{Creative Generation}}} & Imagination & Imagination & Does the image demonstrate creative originality and imaginative thinking when combining novel or surreal elements? \\
             & Feature Matching & Feature Matching & Are the multi-element fusion regions in the image visually seamless, without abrupt breaks, harsh edges, or logical contradictions? \\
             & Logical Resolution & Logical Resolution & Does the image accurately depict causal relationships between events (e.g., breaking glass $\rightarrow$ shards flying, rain $\rightarrow$ wet surfaces)? \\
             & Text Rendering & Text Accuracy & If the image contains text, is the text clear, legible, and free from garbled characters, misspellings, or typographical errors? \\
             &  & Text Layout & Is the text layout (e.g., centering, alignment, line spacing, margins) in the image visually appealing and professionally structured? \\
             &  & Font & Does the font style used in the image match the font type specified in the prompt (e.g., SimSun, Heiti, handwritten, serif)? \\
             &  & Cross-lingual Generation & Does the image correctly follow the translation instructions in the prompt, producing accurate text in the target language? \\
             & Design Applications & Graphic Design & Does the graphic design (e.g., advertisement, poster) exhibit a clear information hierarchy, effective visual guidance, and professional layout? \\
             &  & Product Design & Does the product design in the image demonstrate reasonable industrial design logic (e.g., ergonomic grip, logical interface placement, structural integrity)? \\
             &  & Spatial Design & Does the interior or architectural space conform to the principles of perspective, proportion, and building design standards? \\
             &  & Fashion Styling & Does the clothing cut and silhouette match the style described in the prompt (e.g., Hanfu, cyberpunk, haute couture)? Does the makeup style (e.g., smoky eyes, nude makeup, theatrical look) suit the occasion and character setting? \\
             &  & Game Design & Do the game props and UI elements have practical in-game usability (e.g., icon recognizability, interactive affordances, clear feedback cues)? \\
             &  & Art Design & Does the image successfully demonstrate the specific artistic design style required by the prompt (e.g., unique brushstrokes, distinctive color scheme, coherent artistic language)? \\
             & Visual Storytelling & Cinematic Style & Does the image reproduce the signature visual language of the specific director referenced in the prompt (e.g., Wes Anderson's symmetrical composition, Wong Kar-wai's warm color palette)? \\
             &  & Camera / Lens Style & Does the image reflect the characteristic imaging effects of the specific photographic equipment or lens referenced in the prompt (e.g., film grain, bokeh, digital sharpening)? \\
             &  & Storyboard Creation & Does the image's scene composition follow the panel layout requirements outlined in the prompt (e.g., three-panel, four-panel, split-screen)? \\
             &  & Shot Sizes & Does the image meet the framing and shot size requirements specified in the prompt (e.g., close-up, medium shot, wide shot)? \\
             &  & Composition & Does the image follow the specific composition rules required by the prompt (e.g., rule of thirds, golden ratio, leading lines)? \\
             &  & Angles & Does the camera angle comply with the prompt's specification (e.g., bird's-eye view, low angle, Dutch angle)? \\
             &  & Comic Creation & Does the image conform to the comic style required by the prompt (e.g., American comics, Japanese manga, European BD)? \\
            \bottomrule
        \end{tabular}
    }
\end{table}
\clearpage